\documentclass[sigconf]{acmart}

\usepackage{bm}
\usepackage[figurename=Fig.]{caption}
\usepackage[labelfont={rm}, textfont={rm}]{subcaption}
\usepackage{multirow}
\usepackage{colortbl}
\usepackage{hhline}

\hypersetup{colorlinks=true,urlcolor=blue}
\setcopyright{none}
\acmDOI{10.1145/3132847.3132889}
\acmISBN{978-1-4503-4918-5/17/11}
\acmConference{CIKM'17}{November 6--10}{Singapore}
\acmYear{2017}
\copyrightyear{2017}
\acmPrice{15.00}

\begin{document}
\title[Joint Topic-Semantic-aware Social Recommendation for Online Voting]{Joint Topic-Semantic-aware Social Recommendation\\for Online Voting}

\author{Hongwei Wang$^{1,2}$, Jia Wang$^2$, Miao Zhao$^2$, Jiannong Cao$^2$, Minyi Guo$^1$}
\authornote{M. Guo is the corresponding author.}
\affiliation{$^1$Shanghai Jiao Tong University, $^2$The Hong Kong Polytechnic University}
\email{wanghongwei55@gmail.com, {csjiawang,csmiaozhao,csjcao}@comp.polyu.edu.hk, guo-my@cs.sjtu.edu.cn}

%
%
%
%

\renewcommand{\shortauthors}{H. Wang et al.}

\begin{abstract}
	Online voting is an emerging feature in social networks, in which users can express their attitudes toward various issues and show their unique interest.
	Online voting imposes new challenges on recommendation, because the propagation of votings heavily depends on the structure of social networks as well as the content of votings.
	In this paper, we investigate how to utilize these two factors in a comprehensive manner when doing voting recommendation.
	First, due to the fact that existing text mining methods such as topic model and semantic model cannot well process the content of votings that is typically short and ambiguous, we propose a novel Topic-Enhanced Word Embedding (TEWE) method to learn word and document representation by jointly considering their topics and semantics.
	Then we propose our Joint Topic-Semantic-aware social Matrix Factorization (JTS-MF) model for voting recommendation.
	JTS-MF model calculates similarity among users and votings by combining their TEWE representation and structural information of social networks, and preserves this topic-semantic-social similarity during matrix factorization.
	To evaluate the performance of TEWE representation and JTS-MF model, we conduct extensive experiments on real online voting dataset.
	The results prove the efficacy of our approach against several state-of-the-art baselines.
\end{abstract}


\maketitle

\section{Introduction}
\begin{figure}
		\includegraphics[width=0.38\textwidth]{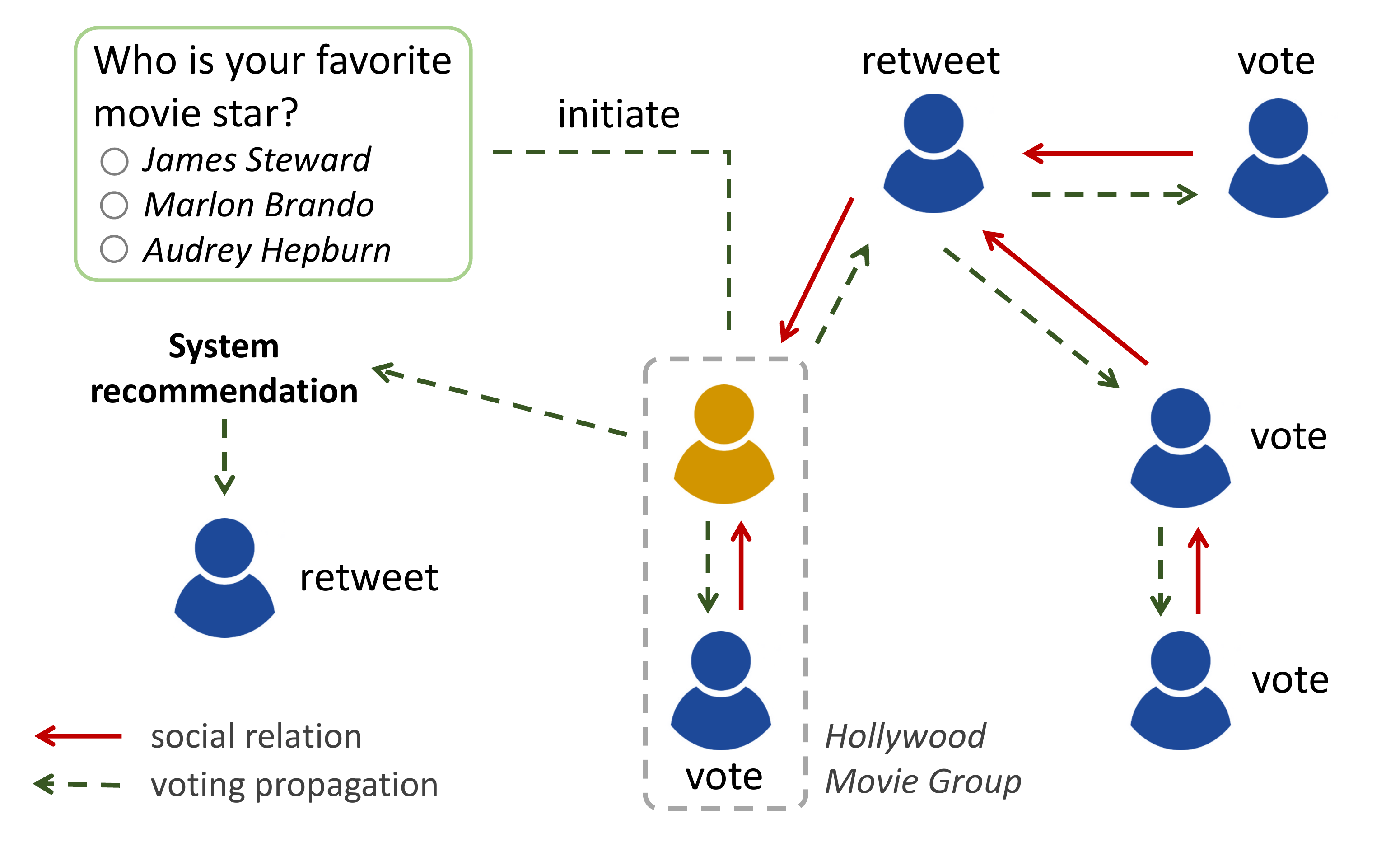}
		\caption{Propagation scheme of online voting.}
		\label{voting_propagation}
		\vspace{-0.15in}
	\end{figure}

	Online voting \cite{xiwang2017voting} has recently become a popular function on social platforms, through which a user can share his opinion towards various interested subjects, ranging from livelihood issues to entertainment news.
	More advanced than simple like-dislike type of votings, some social networks, such as Weibo\footnote{\url{http://www.weibo.com}.}, have empowered users to run their own voting campaigns.
	Users can freely initiate votings on any topics of their own interests and customize voting options.
	These votings are visible to the friends of initiator, who can then choose to participate to make the votings further seen by their friends or simply retweet the votings to their friends. In such a way, in addition to the system recommendation, a voting can widely propagate over the network along social paths.
	The voting propagation scheme is shown in Fig. \ref{voting_propagation}.
	
	Facing a large volume of diversified votings, a critical challenge is to present ``right'' votings to the ``right'' person.
	An effective recommender system is desired to be able to deal with information overload \cite{bobadilla2013recommender} by precisely locating what votings favor each user most, thus improves user experience and maximizes user engagement in votings.
	Such a recommender system can also benefit a variety of other online services such as personalized advertising \cite{tang2016optimizing}, market research \cite{ilieva2002online}, public opinion analysis \cite{liu2012survey}, etc.
	
	Despite the great importance, there is little prior work considering recommending votings to users in social networks.
	The challenges are two-fold.
	First, the propagation of online votings relies heavily on the structure of social networks.
	A user can see the votings initiated, participated or retweeted by his followees, which implies that the user is more likely to be exposed to the votings that his friends are involved in.
	Moreover, in most social networks, a user can join different interest groups, which is another type of social structure that potentially affects users' voting behavior.
	Though several prior works \cite{shi2012climf, gao2015content, jiang2015social, bressan2016limits, hung2016social, zhang2016inferring, wang2016social, xiwang2017voting} have been proposed to leverage social network information in recommendation, it is still an open question how to comprehensively incorporate structural social information into the task of voting recommendation considering its propagation pattern.
	Second, users' interest in votings is strongly connected with voting content presented in question text (e.g., ``Who is your favorite movie star?'').
	Topic model \cite{Blei2003LDA} is regarded as a possible approach to mine the voting interests through discovering the latent topic distribution of relevant voting text.
	However, the voting questions are typically short and lack sufficient topic information, leading to severe performance degradation of topic models.
	Alternatively, semantic analytics \cite{Mikolov2013wordembedding} can also possibly be used to mine voting interests through learning text representations.
	However, such semantic models typically represent each word using a single vector, making them indiscriminative for homonymy and polysemy, which are especially common in voting questions (e.g., ``Do you use apple products?'' and ``Do you peel apple before eating?'').
	In brief, these inherent defects of the above models limit their power in the scenario of social voting recommendation. 
	
	To address aforementioned challenges, in this paper, we propose a novel \textit{Joint Topic-Semantic-aware Matrix Factorization} (JTS-MF) model for online voting recommendation.
	JTS-MF model considers social network structure and representation of voting content in a comprehensive manner.
	For social network structure, JTS-MF model fully encodes the information of social relationship and group affiliation into the objective function.
	We will further justify the usage of social network structure in Section \ref{background_and_data_analysis}.
	For representation of voting content, we propose a \textit{Topic-Enhanced Word Embedding} (TEWE) method to build a multi-prototype word and document\footnote{In this paper, a ``document''  can be related to a voting, a user or a group. A voting document is the content of its question, a user document is formed by aggregating all the documents of votings he participates, and a group document is formed by aggregating all the documents of users who join the group.} representation, which jointly considers their topics and semantics.
	The key idea of TEWE is to enable each word to have different representations under different word topics and different documents.
	We will detail TEWE in Section \ref{joint_topic_semantic_embedding_learning}.
	Once obtaining TEWE representation for each document, JTS-MF model combines them with the structural information of social networks to calculate the topic-semantic-social similarity among users and votings.
	The reason of calculating such similarity is that, inspired by Locally Linear Embedding \cite{roweis2000nonlinear}, we try to preserve the similarity among users and votings during matrix factorization, as it contains abundant proximity information and can greatly benefit feature learning for users and votings.
	JTS-MF model is detailed in Section \ref{recommendation_model}.
	
	We conduct extensive investigation on JTS-MF with real online voting dataset.
	The experimental results in Section \ref{experiments} demonstrate that JTS-MF model achieves substantial gains compared with baselines.
	The results also prove that TEWE is able to well combine topic and semantic information of texts and generates a better kind of document representation.
	
	In summary, the contributions of this paper are as follows:
	\begin{itemize}
		\item
			We formally formulate the online voting recommendation problem, which has not been fully investigated yet.
		\item We indicate that user's voting behavior is highly correlated with social network structure by conducting thorough statistical measurements.
		\item
			We propose a novel Topic-Enhanced Word Embedding model to jointly consider topics and semantics of words and documents to learn their representation.
			TEWE takes advantages of both topic model and semantic model, and consequently learns more informative embeddings.
		\item
			We develop a novel matrix factorization based models, named JTS-MF, for online voting recommendation.
			JTS-MF is able to preserve the topic-semantic-social similarity among users and votings from original embedding space during learning process.
		\item
			We carry out extensive experiments on real online voting dataset, the results of which reveal that JTS-MF significantly outperforms baseline (variant) methods, say for example, surpassing basic matrix factorization model with $57\%$, $38\%$ and $25\%$ enhancement in terms of recall for top-$1$, top-$5$ and top-$20$ recommendation, respectively.
	\end{itemize}

\section{Related Work}
	\subsection{Recommender Systems}
		Roughly speaking, existing recommender systems can be categorized into three classes \cite{bobadilla2013recommender}: content-based, collaborative filtering, and hybrid methods.
		Content-based methods \cite{lang1995newsweeder, zhu2012mining} make use of user profiles or item descriptions as features for recommendation.
		Collaborative filtering methods \cite{shi2012climf, rendle2014improving, wang2016recommending, xiwang2017voting} use either explicit feedback (e.g., users' ratings on items) or implicit feedback (e.g., users' browsing records about items) data of user-item interactions to find user preference and make the recommendation.
		In addition, various models are incorporated into collaborative filtering, such as Support Vector Machine \cite{xia2006support}, Restricted Boltzmann Machine \cite{salakhutdinov2007restricted}, and Stacked Denoising Auto Encoder \cite{wang2015collaborative}.
		Hybrid methods \cite{li2011generalized, hu2013personalized} combine content-based and collaborative filtering models in many hybridization approaches, such as weighted, switching, cascade and feature combination or augmentation.
			
	\subsection{Social Recommendation}
		Traditional recommender systems are vulnerable to data sparsity problem and cold-start problem.
		To mitigate this issue, many approaches have been proposed to utilize social network information in recommender systems \cite{shi2012climf, gao2015content, jiang2015social, bressan2016limits, hung2016social, zhang2016inferring, wang2016social, xiwang2017voting}.
		For example, \cite{jiang2015social} represents a social network as a star-structured hybrid graph centered on a social domain which connects with other item domains to help improve the prediction accuracy.
		\cite{hung2016social} investigates the seed selection problem for viral marketing that considers both effects of social influence and item inference for product recommendation.
		\cite{wang2016social} studies the effects of strong and weak ties in social recommendation, and extends Bayesian Personalized Ranking model to incorporate the distinction of strong and weak ties.
		However, the above works only utilize users' social links without considering the topic and semantic information for mining the similarities among users and items, which we found quite helpful for voting recommendation tasks.
		Another difference between these works and ours is that we also take social group affiliation into consideration, which can further improve the performance of recommendation.
	
	\subsection{Topic and Semantic Language Models}
		Latent Dirichlet Allocation (LDA) \cite{Blei2003LDA} is a well-known generative topic model that learns the latent topic distributions for documents.
		LDA is widely used in sentiment analysis \cite{mei2007topic}, aspects and opinions mining \cite{zhao2010jointly}, and recommendation \cite{diao2014jointly}.
		Word2vec \cite{Mikolov2013wordembedding} is generally recognized as a neural network model, which learns word representations that capture precise syntactic and semantic word relationships.
		Word2vec as well as associated Skip-Gram model are extensively used in document classification \cite{le2014distributed}, dependency parser \cite{chen2014fast}, and network embedding \cite{grover2016node2vec}.
		However, LDA and Word2vec are not directly applicable in the scenario of voting recommendation because the content of voting is usually short and ambiguous.
		As a combination, \cite{liu2015topical} tries to learn topical word embeddings based on both words and their topics.
		The difference between \cite{liu2015topical} and ours is that we also take topics of documents into consideration, which enables our model to learn a even more discriminative and informative representations for words and documents.

\section{Background and Data Analysis}
\label{background_and_data_analysis}
	In this section, we briefly introduce the background of Weibo voting and present detailed analysis of Weibo voting dataset.
	
	\subsection{Background}
		Weibo is one of the most popular Chinese microblogging website launched by Sina corporation, which is akin to a hybrid of Twitter and Facebook platforms.
		Users on Weibo can follow each other, write tweets and share with his followers.
		Users can also join different groups based on their attributes (e.g., region) or interests of topics (e.g., career).
		
		Voting\footnote{\url{http://www.weibo.com/vote?is_all=1}.} is one of the embedded features of Weibo.
		As of January 2013, more than 92 million users have participated in at least one voting and more than 2.2 million ongoing votings are available on Weibo every day.
		Any user can freely initiate, retweet and participate a voting campaign in Weibo.
		As shown in Fig. \ref{voting_propagation}, votings can propagate in two ways.
		The first way is through social propagation: a user can see the voting initiated, retweeted or participated by his followees and potentially participates the voting.
		The second way is through Weibo voting recommendation list, which consists of popular votings and personalized recommendation for each user.

	\subsection{Data Measurements}
		Our Weibo voting dataset comes from the technical team of Sina Weibo, which contains detailed information about votings from November 2010 to January 2012, as well as other auxiliary information.
		Specifically, the dataset includes users' participation status on each voting\footnote{We only know whether a user participated a voting or not, rather than user voting results, i.e., we do not know which voting option a user chose.}, content of each voting, social connection between users, name and category of each group, and user-group affiliation.
		
		\subsubsection{Basic statistics.}
\begin{figure}
			\centering
            		\begin{subfigure}[b]{0.23\textwidth}
                		\includegraphics[width=\textwidth]{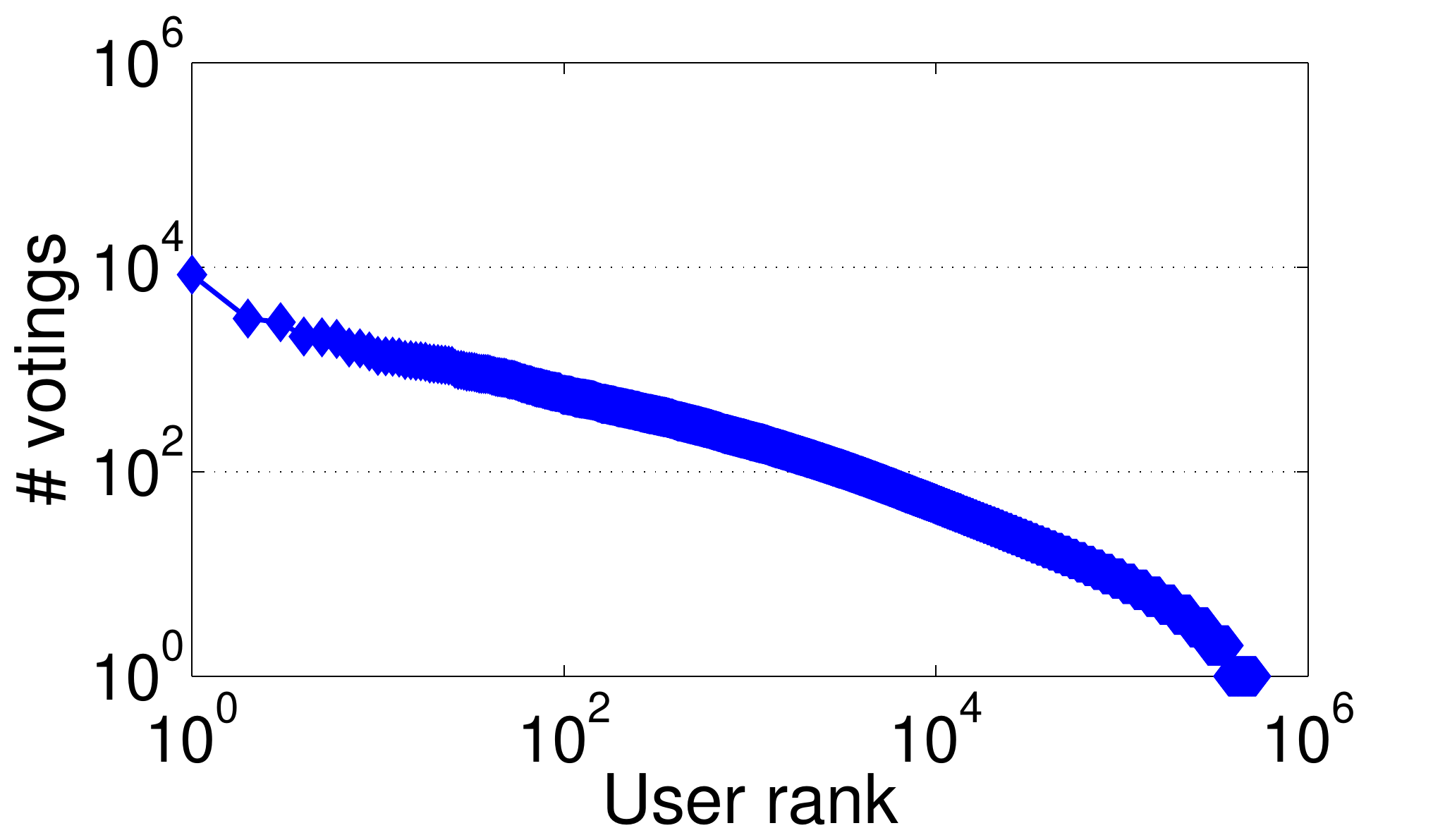}
                		\caption{}
                		\label{fig:statistics_a}
            		\end{subfigure}
            		\hfill
            		\begin{subfigure}[b]{0.23\textwidth}
                		\includegraphics[width=\textwidth]{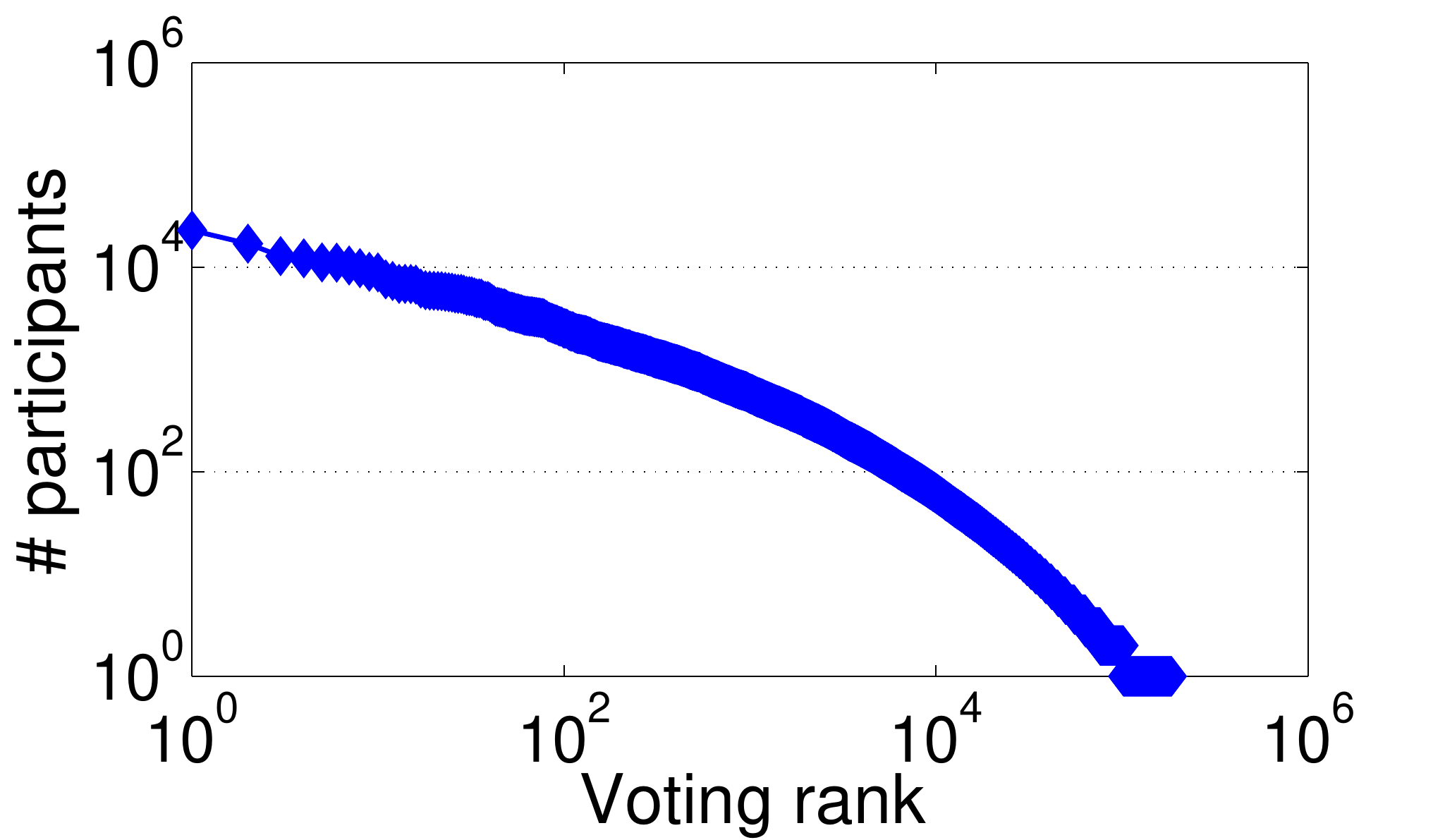}
                		\caption{}
                		\label{fig:statistics_b}
            		\end{subfigure}
            		\hfill
            		\begin{subfigure}[b]{0.23\textwidth}
                		\includegraphics[width=\textwidth]{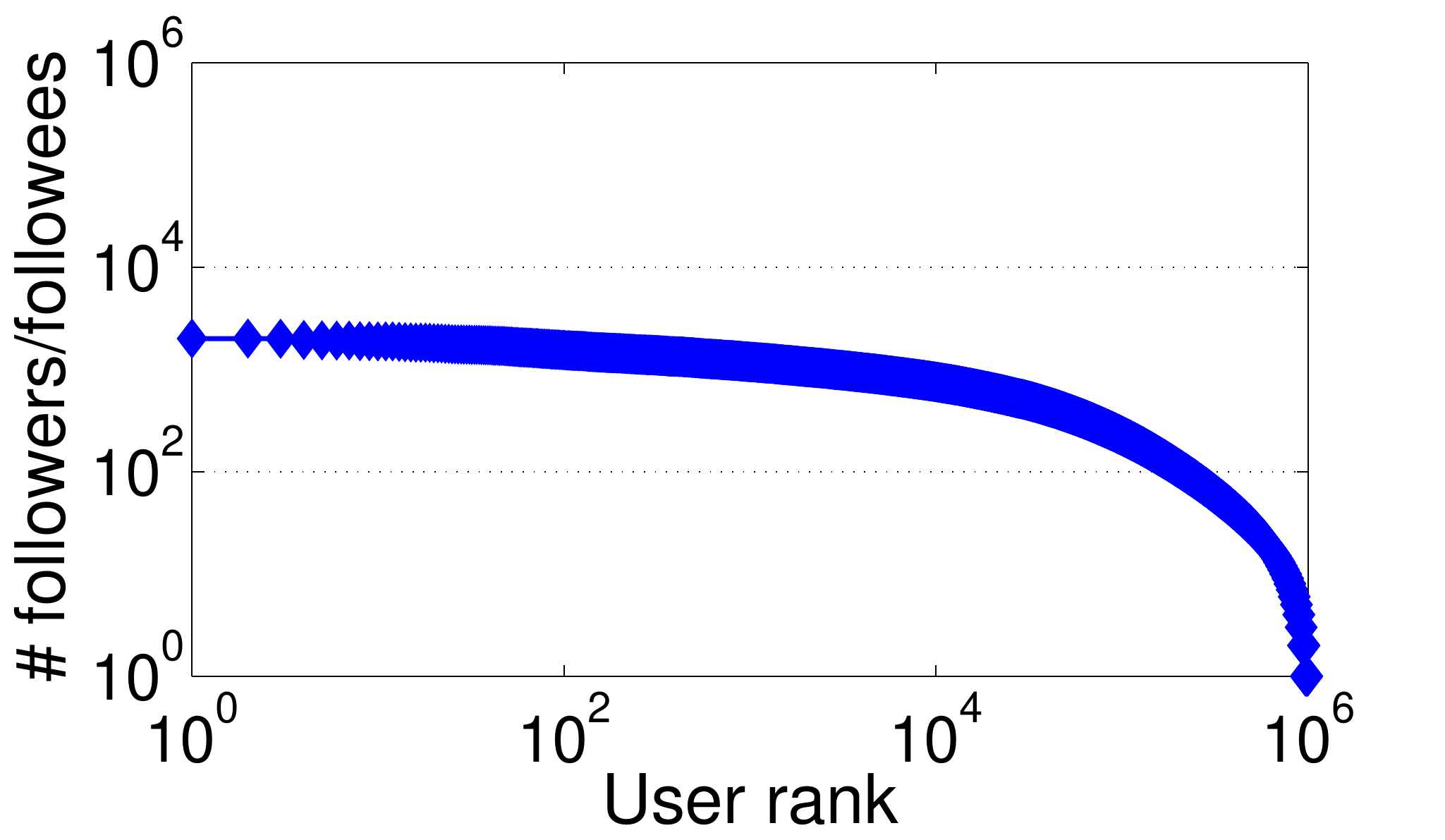}
                		\caption{}
                		\label{fig:statistics_c}
            		\end{subfigure}
        		\hfill
            		\begin{subfigure}[b]{0.23\textwidth}
                		\includegraphics[width=\textwidth]{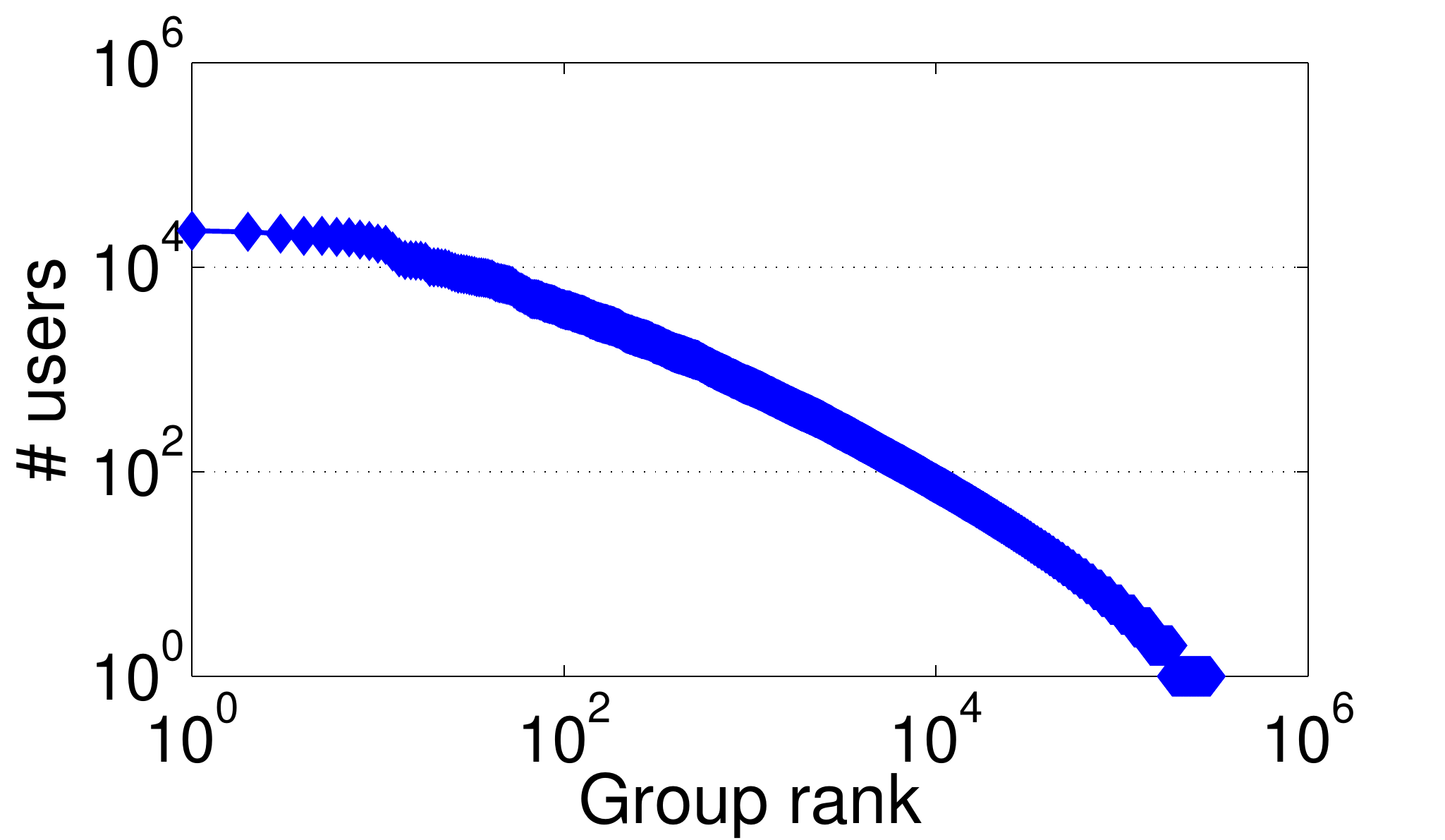}
                		\caption{}
                		\label{fig:statistics_d}
            		\end{subfigure}
            		\hfill
            		\begin{subfigure}[b]{0.23\textwidth}
                		\includegraphics[width=\textwidth]{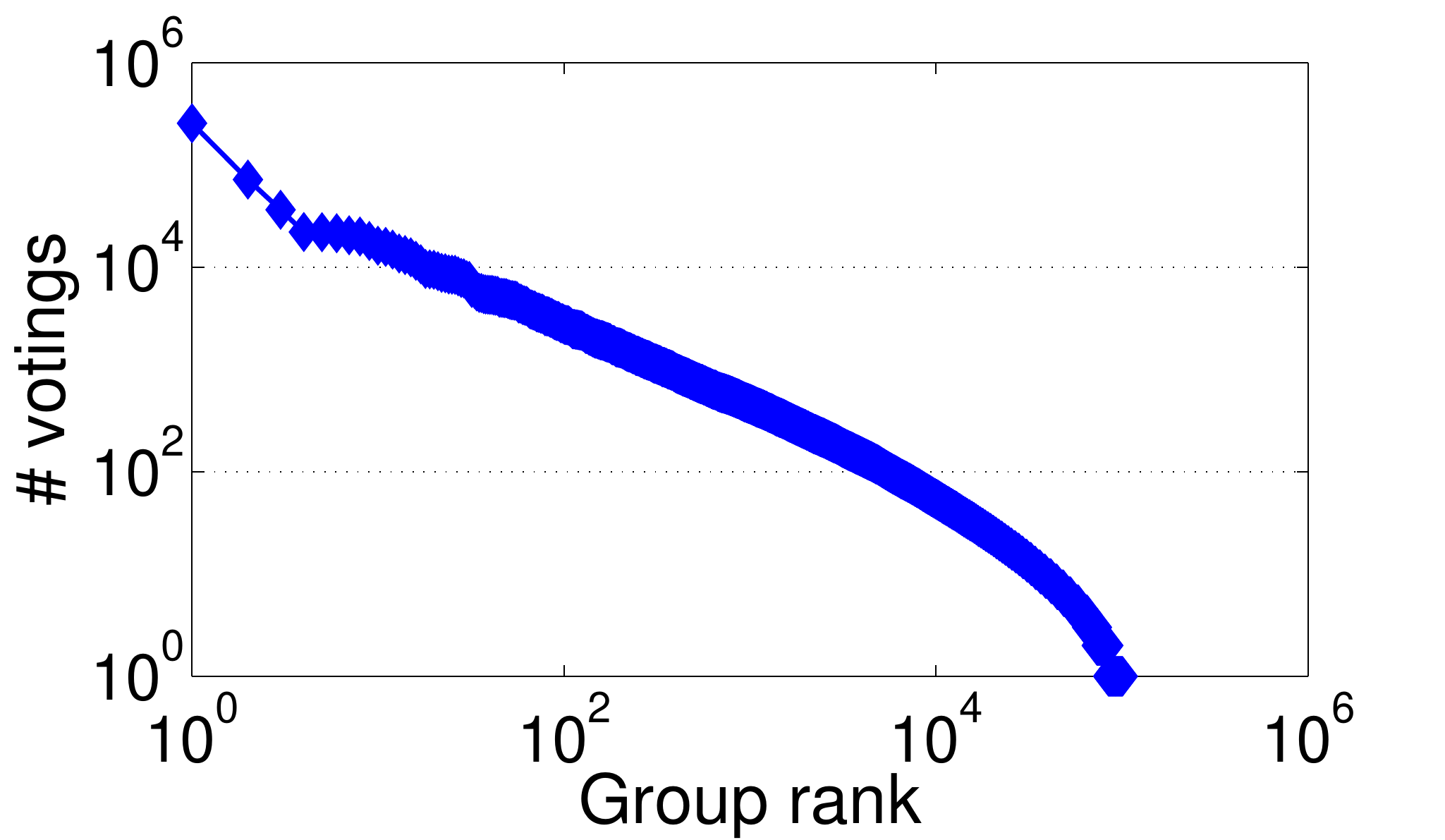}
                		\caption{}
                		\label{fig:statistics_e}
            		\end{subfigure}
            		\hfill
            		\begin{subfigure}[b]{0.23\textwidth}
                		\includegraphics[width=\textwidth]{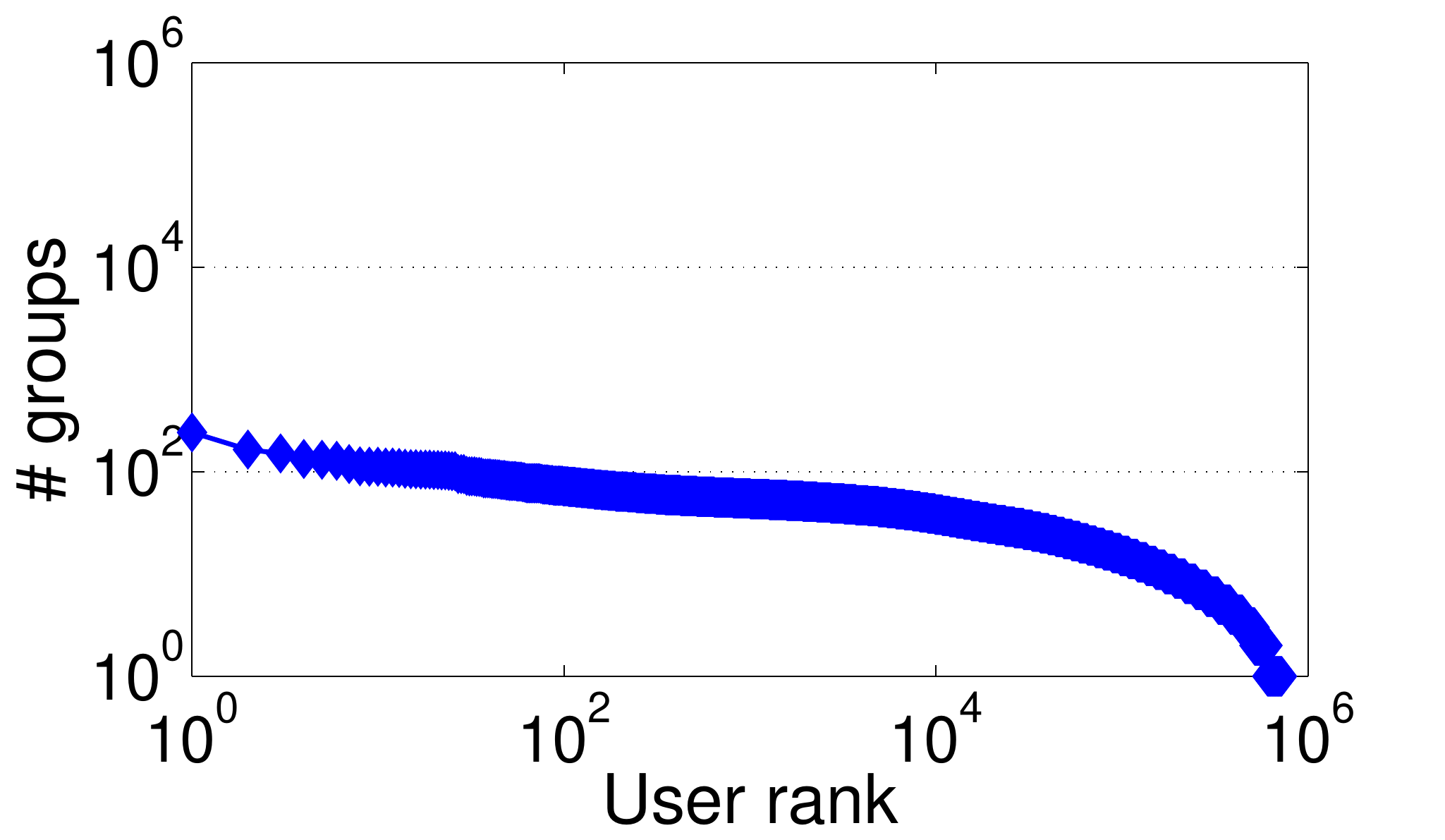}
                		\caption{}
                		\label{fig:statistics_f}
            		\end{subfigure}
            		\vspace{0.1in}
            		\caption{(a) Distribution of the number of votings participated by a user; (b) Distribution of the number of participants of a voting; (c) Distribution of the number of followers/followees of a user; (d) Distribution of the number of users in a group; (e) Distribution of the number of votings (may contain duplicated votings) participated by all users in a group; (f) Distribution of the number of groups joined by a user.}
            		\label{fig:statistics}
            		\vspace{-0.15in}
        	\end{figure}

		The basic statistics are summarized in Table \ref{table:statistics}.
		From Table \ref{table:statistics} we can learn that, each user has $165.4$ followers/followees, participates $3.9$ votings, and joins $5.6$ groups on average.
		If we only count users who participate at least one voting and users who join at least on group, the average number of votings and average number of joined groups of each user is $7.4$ and $7.8$, respectively.
		Fig. \ref{fig:statistics} depicts the distribution curves of the above statistics, where the meaning of each subfigure is given in the caption.
		
		\begin{table}
			\centering
			\small
			\caption{Basic Statistics of Weibo Dataset.}
			\begin{tabular}{|l|r||l|r|}
				\hline
				\# users & 1,011,389 & \# groups & 299,077\\
				\hline
				\# users with votings & 525,589 & \# user-voting & 3,908,024\\
				\hline
				 \# users with groups & 723,913 & \# user-user & 83,636,677\\
				\hline
				\# votings & 185,387  & \# user-group & 5,643,534\\
				\hline
			\end{tabular}
			\label{table:statistics}\vspace{-0.1in}
		\end{table}

        	To get an intuitive understanding of whether user's voting behavior is correlated with his social relation and group affiliation, we conduct the following two sets of statistical experiments:

		\subsubsection{Correlation between the number of common votings of user pairs and the types of user pairs.}
			We randomly select ten million user pairs from the set of all users, and count the average number of votings that the two users both participate under the following four circumstances:
			1) one of the users follows the other in the pair, i.e., they are social-level friends;
			2) the two users are in at least one common group, i.e., they are group-leven friends;
			3) the two users are neither social-level friends nor group-level friends;
			4) all cases.
			The results are plotted in Fig. \ref{fig:cor1}, which clearly shows the difference among these cases.
			In fact, the average number of common votings of social-level friends ($3.54 \times 10^{-4}$) and group level friends ($1.79 \times 10^{-4}$) are 17.4 and 8.8 times higher than that of ``strangers'' ($2.04 \times 10^{-5}$).
			The results demonstrate that if two users are social-level or group-level friends, they are likely to participate more votings in common.

		\subsubsection{Correlation between the probability of two users being friends and whether they participate common voting.}
\begin{figure}
				\centering
				\begin{subfigure}[b]{0.23\textwidth}
   					\includegraphics[width=\textwidth]{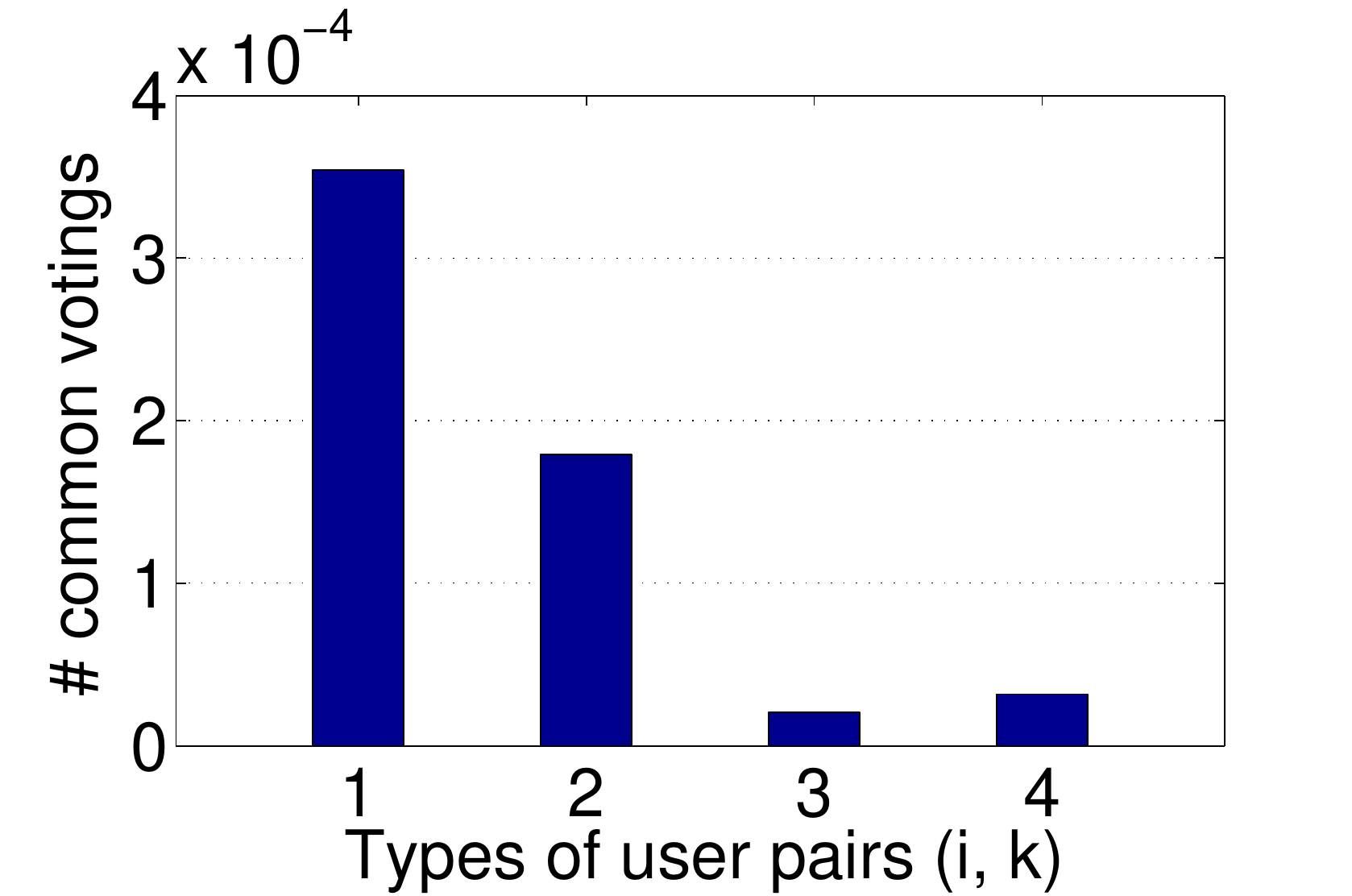}
   					\caption{}
   					\label{fig:cor1}
				\end{subfigure}
				\hfill
				\hspace{-0.1in}
				\begin{subfigure}[b]{0.23\textwidth}
					\includegraphics[width=\textwidth]{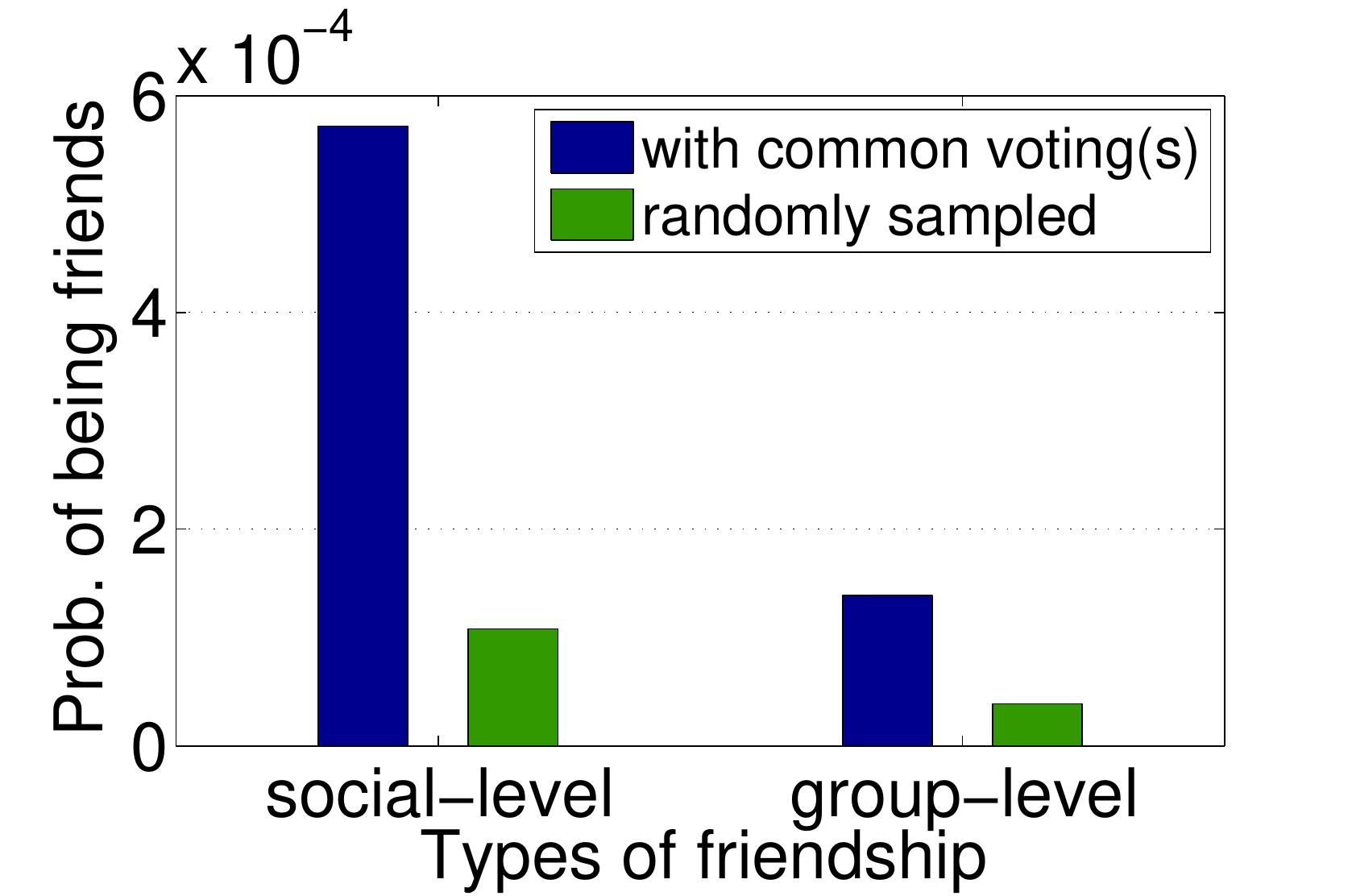}
					\caption{}
					\label{fig:cor2}
				\end{subfigure}
				\vspace{0.05in}
				\caption{(a) Average number of common votings participated by user $u_i$ and $u_k$ in four cases: 1. $u_i$ follows/is followed by $u_k$; 2. $u_i$ and $u_k$ are in at least one common group; 3. $u_i$ and $u_k$ have no social-level and group-level relationship; 4. all cases; (b) Probability of two users being social-level or group-level friends in two cases: 1. they ever participated at least one common voting; 2. they are randomly sampled.}
				\vspace{-0.05in}
			\end{figure}

			We first randomly select ten thousand votings from the set of all votings.
			For each sampled voting $v_j$, we calculate the probability that two of its participants are social-level or group-level friends, i.e., $p_j = \frac{\text{\# of social/group-level friends among participants of } v_j}{n_j \times (n_j - 1) / 2}$, where $n_j$ is the number of $v_j$' participants.
			We calculate $p_j$ over all sampled votings and plot the average result (blue bar) in Fig. \ref{fig:cor2}.
			For comparison, we also plot the result for randomly sampled set of users (green bar) in Fig. \ref{fig:cor2}.
			It is clear that if two users ever participated common voting, they are more likely to be social-level or group-level friends.
			In fact, probabilities of two users being social-level or group-level friends are raised by 5.3 and 3.6 times given the observation that they are with common voting.
				
			The above two findings effectively prove the strong correlation between voting behavior and social network structure, which motivates us to take users' social relation and group affiliation into consideration when making voting recommendation.	
		
\section{Problem Formulation}
\label{problem_formulation}
	In this paper, we consider the problem of recommending Weibo votings to users.
	We denote the set of all users, the set of all votings, and the set of all groups by $\mathcal U=\{u_1, ..., u_N\}$, $\mathcal V=\{v_1, ..., v_M\}$, and $\mathcal G=\{G_1, ..., G_L\}$, respectively.
	Moreover, we model three types of relationship in Weibo platform: user-voting, user-user, and user-group relationship as follows:
		
	\begin{enumerate}
		\item
			The user-voting relationship for $u_i$ and $v_j$ is defined as
			\begin{equation}
			\label{eq:uv_relation}
				I_{u_i, v_j}=
					\begin{cases}
						1, & if \ u_i \ participates \ v_j;\\
						0, & otherwise.
					\end{cases}
			\end{equation}
		\item
			The user-user relationship for $u_i$ and $u_k$ is defined as
			\begin{equation}
			\label{eq:uu_relation}
				I_{u_i, u_k}=
					\begin{cases}
						1, & if \ u_i \ follows \ u_k;\\
						0, & otherwise.
					\end{cases}
			\end{equation}
			We further use $\mathcal F^+_i$ to denote the set of $u_i$'s followees, and use $\mathcal F^-_i$ to denote the set of $u_i$'s followers (``$+$'' means ``out'' and ``$-$'' means ``in'').
		\item
			The user-group relationship for $u_i$ and $G_c$ is defined as
			\begin{equation}
			\label{eq:ug_relation}
				I_{u_i, G_c}=
					\begin{cases}
						1, & if \ u_i \ joins \ G_c;\\
						0, & otherwise.
					\end{cases}
			\end{equation}
	\end{enumerate}
		
	Given the above sets of users and votings as well as three types of relationship, we aim to recommend a list of votings for each user, in which the votings are not participated by the user but may be interesting to him.

\section{Joint-topic-semantic Embedding}
\label{joint_topic_semantic_embedding_learning}
	In this section, we explain how to learn the embeddings of users, votings, and groups in a joint topic and semantic way, and apply the embeddings to calculate similarities.
	We first introduce the methods of learning topic information and semantic information by LDA and Skip-Gram models, respectively, and propose our method which combines these two models to learn more powerful embeddings.
		
	\subsection{Topic Distillation}
	\label{section:lda}
		In this subsection, we introduce how to profile users, votings, and groups in terms of topic interest distribution by performing topic distillation on the associated textual content information.
		
		In general, LDA is a popular generative model to discover latent topic information from a collection of documents \cite{Blei2003LDA}.
		In LDA, each document $d$ is represented as a multinomial distribution $\bm \Theta_d$ over a set of topics, and each topic $z$ is also represented as a multinomial distribution $\bm \Phi_z$ over a number of words.
		Subsequently, each word position $l$ in document $d$ is assigned a topic $z_{d, l}$ according to $\bm \Theta_d$, and the word $w_{d, l}$ is finally generated according to $\bm \Phi_{z_{d, l}}$.
		By LDA approach, the topic distribution for each document and the topic assignment for each word can be obtained, which would be utilized later in our proposed model.
		
		Here, we discuss how to apply LDA in the scenario of Weibo voting.
		According to the Weibo voting dataset, each voting $v_j$ associates a sentence of question, which can be regarded as document $d_{v_j}$\footnote{$d_{v_j}$ is segmented by Jieba (\url{https://github.com/fxsjy/jieba}) and all stop words are removed.}.
		The document $d_{u_i}$ for user $u_i$ can thus be formed by aggregating the content of all votings he participates, i.e., $d_{u_i} = \cup \{ d_{v_j} | I_{u_i, v_j} = 1 \}$, and the document $d_{G_c}$ for group $G_c$ is formed by aggregating documents of all its members, i.e., $d_{G_c} = \cup \{ d_{u_i} | I_{u_i, G_c} = 1 \}$.
		Note that though our target is to learn the topic distributions of all users, votings, and groups, it is inadvisable to train LDA model on $d_{u_i}$'s and $d_{v_j}$'s because:
		(1) the entitled sentence associated with a single voting is typically short-presented and topic-ambiguous;
		(2) even with user-level voting content aggregation, some documents of inactive users are not long enough to accurately extract the authentic topic distribution, yet showing relatively flat distribution over all the topics.
		Therefore, we choose to feed group-level aggregated documents $d_{G_c}$'s to LDA model as training samples.
		The process of group-level voting content aggregation will cover all the content the affiliated users are interested in and help better identify their interests in terms of voting topic.

		Denote $\text{Dir}(\bm \alpha)$ as the Dirichlet prior of $\bm \Theta_d$, and $\text{Dir}(\bm \beta)$ as the Dirichlet prior of $\bm \Phi_z$.
		Given $\bm \alpha$ and $\bm \beta$, the joint distribution of document-topic distributions $\bm \Theta$, topic-word distributions $\bm \Phi$, topics of words $\bm z$, and a set of words $\bm w$ is
		\begin{equation}
		\label{lda_likelihood}
			\begin{array}{l}
				p (\bm \Theta, \bm \Phi, \bm z, \bm w | \bm \alpha, \bm \beta)\\
				=\prod_z p(\bm \Phi_z | \bm \beta) \cdot \prod_d \bigg( p(\bm \Theta_d | \bm \alpha) \prod_l \Big( p(z_{d, l} | \bm \Theta_d) p(w_{d, l} | \bm \Phi_{z_{d, l}}) \Big) \bigg),
			\end{array}
		\end{equation}
		where $d$ traverses all group-level aggregated documents.
		In general, it is computationally intractable to directly maximize the joint likelihood in Eq. (\ref{lda_likelihood}), thus Gibbs Sampling \cite{Griffiths2004Gibbssampling} is usually applied to estimate the posterior probability $p(\bm z | \bm w, \bm \alpha, \bm \beta)$ and solve $\bm \Theta$, $\bm \Phi$ iteratively.
		Denote $\theta_d^{(z)}$ the $z$-th component of $\bm \Theta_d$, and $\phi_z^{(w)}$ the $w$-th component of $\bm \Phi_z$.
		With the sampling results, $\bm \Theta_d$ and $\bm \Phi_z$ can be estimated as:		
		\begin{equation}
		\label{eq:gibbs}
			\begin{array}{ll}
				\theta_d^{(z)} = \big( n_d^{(z)} + \alpha^{(z)} \big) / \big( \sum_z (n_d^{(z)} + \alpha^{(z)} ) \big), &z = 1, ..., Z,\\
				\phi_z^{(w)} = \big( n_z^{(w)} + \beta^{(w)} \big) / \big( \sum_w (n_z^{(w)} + \beta^{(w)} ) \big), &w = 1, ..., V,
			\end{array}
		\end{equation}
		where $\alpha^{(z)}$ is the $z$-th component of $\bm \alpha$, $\beta^{(w)}$ is the $w$-th component of $\bm \beta$, $ n_d^{(z)}$ is the observation counts of topic $z$ for document $d$, $ n_z^{(w)}$ is the frequency of word $w$ assigned as topic $z$, $Z$ is the number of topics and $V$ is vocabulary size.
		
		So far, we have obtained the topic assignment for each word and topic distribution for each group.
		Topic distributions for users and votings can thus be inferred by using the learned model and Gibbs Sampling, which is similar to the calculation of $\theta_d^{(z)}$ in Eq. (\ref{eq:gibbs}).

	\subsection{Semantic Distillation}
		In this subsection, we introduce how to profile users, votings, and groups in terms of semantic information.
		Word embedding, which represents each word using a vector, is widely used to capture semantic information of words.
		Skip-Gram model is a well-known framework for word embedding, which finds word representation that are useful for predicting surrounding words in a document given a target word in a sliding window \cite{Mikolov2013wordembedding}.
		More formally, given a word sequence $D=\{ w_1, w_2, \ldots, w_T \}$, the objective function of Skip-Gram is to maximize the average log probability
		\begin{equation}
			\mathcal{L}(D) = \frac{1}{T} \sum_{t = 1}^T \sum_{\substack{-k \leq c \leq k \\ c \neq 0}} \log p(w_{t+c} | w_t),
		\end{equation}
		where $k$ is the training context size of the target word, which can be a function of the centered work $w_{t}$.
		The basic Skip-Gram formulation defines $p(w_i|w_t)$ using the softmax function as follows:
		\begin{equation}
			p(w_i | w_t)=\frac{\exp ({\bf w}_i^\top {\bf w}_t )}{\sum_{w \in V} \exp ({\bf w}^\top {\bf w}_t )},
		\end{equation}
		where ${\bf w}_i$ and ${\bf w}_t$ are the vector representation of context word $w_i$ and target word $w_t$, respectively, and $V$ is the vocabulary.
		To avoid traversing the entire vocabulary, hierarchical softmax or negative sampling are used in general during learning process \cite{Mikolov2013wordembedding}.

	\subsection{Topic-Enhanced Word Embedding}
	\label{section:tewe}
		In this subsection, we propose a joint topic and semantic learning model, named \textit{Topic-Enhanced Word Embedding} (TEWE), to analyze documents of users, votings, and groups.
		The motivation of proposed TEWE is based on the following two observations:
		(1) The voting content typically involves short texts.
		Even we infer the topic distribution for each voting based on the learned topic-word distribution from group-level aggregated documents, it is still topic-ambiguous to some extent.
		(2) The Skip-Gram model for word embedding assumes that each word always preserves a single vector, which sometimes is indiscriminate under different circumstances due to the homonymy and polysemy.
		Therefore, the basic idea of TEWE is to preserve topic information of documents and words when measuring the interaction between target word $w_t$ and context word $w_i$.
		In this way, a word with different associated topics has different embeddings, and a word in documents with different topics has different embeddings, too.

		Specifically, rather than solely using the target word $w$ to predict context words in Skip-Gram, TEWE also jointly utilizes $z_w$, the topic of the word in a document, as well as $z_w^d$, the most likely topic of the document that the word belongs to.
		Recall that in Section \ref{section:lda}, we have obtained the topic of each word $z_w$ and topic distributions of each document $\bm \Theta_d$, thus $z_w^d$ can be calculated as $z_w^d = \arg \max_z \theta_d^{(z)}$, where $\theta_d^{(z)}$ is the probability that document $d$ belongs to topic $z$, as introduced in Eq. (\ref{eq:gibbs}).	
		TEWE regards each word-topics triplet $\langle w, z_w ,z_w^d \rangle$ as a pseudo word and learns a unique vector ${\bf w}^{z, z^d}$ for it.
		The objective function of TEWE is as follows:
		\begin{equation}\hspace{-0.1in}
			\mathcal{L}(D) = \frac{1}{T} \sum_{t = 1}^T \sum_{\substack{-k \leq c \leq k \\ c \neq 0}} \log p ( \langle w_{t+c}, z_{t+c}, z_{t+c}^d \rangle | \langle w_t, z_t, z_t^d \rangle ),
		\end{equation}
		where $p ( \langle w_i, z_i, z_i^d \rangle | \langle w_t, z_t, z_t^d \rangle )$ is a softmax function as
		\begin{equation}
			p ( \langle w_i, z_i, z_i^d \rangle | \langle w_t, z_t, z_t^d \rangle ) = \frac{\exp \Big( {\bf w}_i^{z, z^d} {}^\top {\bf w}_t^{z, z^d} \Big)}{\sum_{\langle w, z, z^d \rangle \in \langle V, Z, Z \rangle} \exp \Big( {\bf w}^{z, z^d} {}^\top {\bf w}_t^{z, z^d} \Big)}.
		\end{equation}
		
\begin{figure}
			\centering
			\begin{subfigure}[b]{0.22\textwidth}
   				\includegraphics[width=\textwidth]{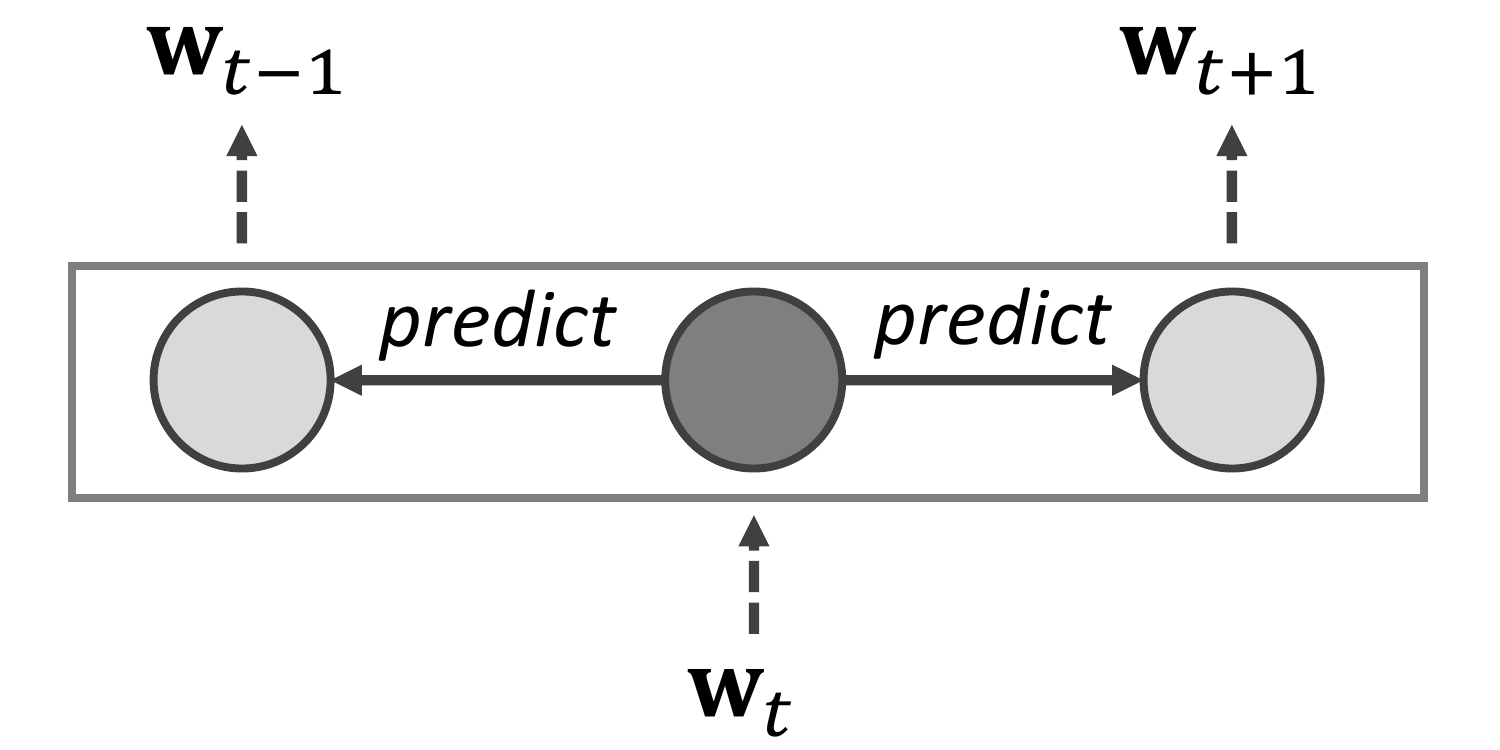}
   				\caption{Skip-Gram}
			\end{subfigure}
			\hfill
			\begin{subfigure}[b]{0.22\textwidth}
				\includegraphics[width=\textwidth]{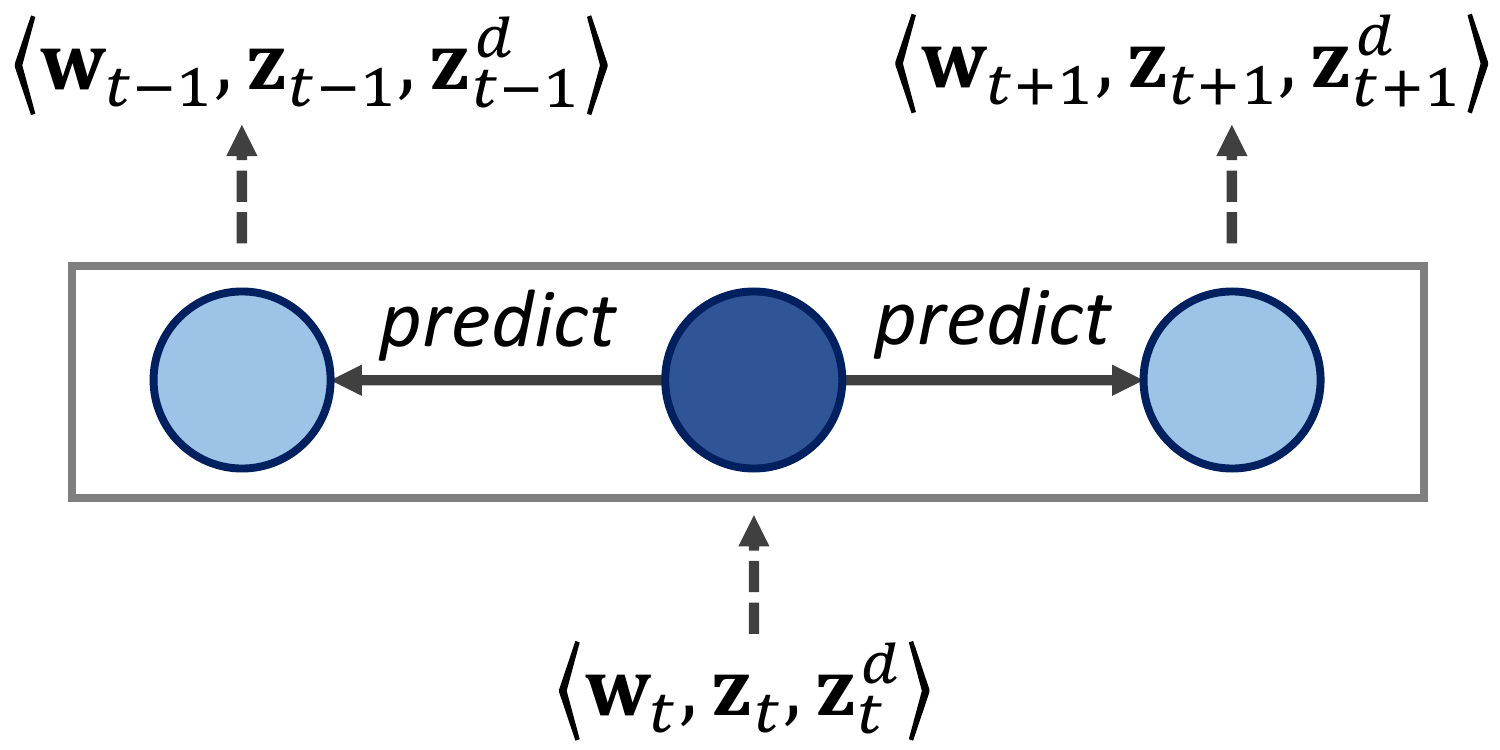}
				\caption{TEWE}
			\end{subfigure}
			\vspace{0.1in}
			\caption{Comparison between Skip-Gram and TEWE. The gray circles in (a) indicate the embeddings of original words, while the blue circles in (b) indicate the TEWE representation of pseudo words, which preserves semantic and topic information of words and documents.}
			\vspace{-0.1in}
			\label{fig:comp}
		\end{figure}

		The comparison between Skip-Gram and TEWE is shown in Fig. \ref{fig:comp}.
		Instead of solely utilizing the target and context words as in Skip-Gram, TEWE further preserves word topic and document topic along with these words, and incorporates both topic and semantic information in embedding learning.
		
		Once obtaining TEWE representation for each pseudo word, the representation of each document can be correspondingly derived by aggregating the embeddings of its containing words weighted by \textit{term frequency-inverse document frequency} (TF-IDF) coefficient.
		Specifically, for each document $d$, its TEWE can be calculated as
		\begin{equation}
		\begin{array}{l}
			{\bf e}_d = \sum_{w \in d} \text{TF-IDF}(w, d) \cdot {\bf w}^{z, z^d},
			\end{array}
		\end{equation}
		where $\text{TF-IDF}(w, d)$ is the product of the raw count of $w$ in $d$ and the logarithmically scaled inverse fraction of the documents that contains $w$, i.e., $\text{TF-IDF}(w, d) = f_{w, d} \cdot \log \frac{|D|}{| d \in D : w \in d |}$ ($D$ is the set of all documents).
		TEWE document representations can be used in measuring inter-document similarities.
		For example, the similarity of two user documents $d_{u_i}$ and $d_{u_k}$ can be calculated as the cosine similarity between their TEWE representations, i.e., $\frac{{\bf e}_{u_i}^{\top}{\bf e}_{u_k}}{\|{\bf e}_{u_{i}}\|_{2}\|{\bf e}_{u_k}\|_{2}}$.
		This similarity encodes both topic and semantic proximity information of user documents, which implicitly reveals the similarity of voting interests between two users.

\section{Recommendation Model}
\label{recommendation_model}
\begin{figure}
		\centering
		\includegraphics[width=.4\textwidth]{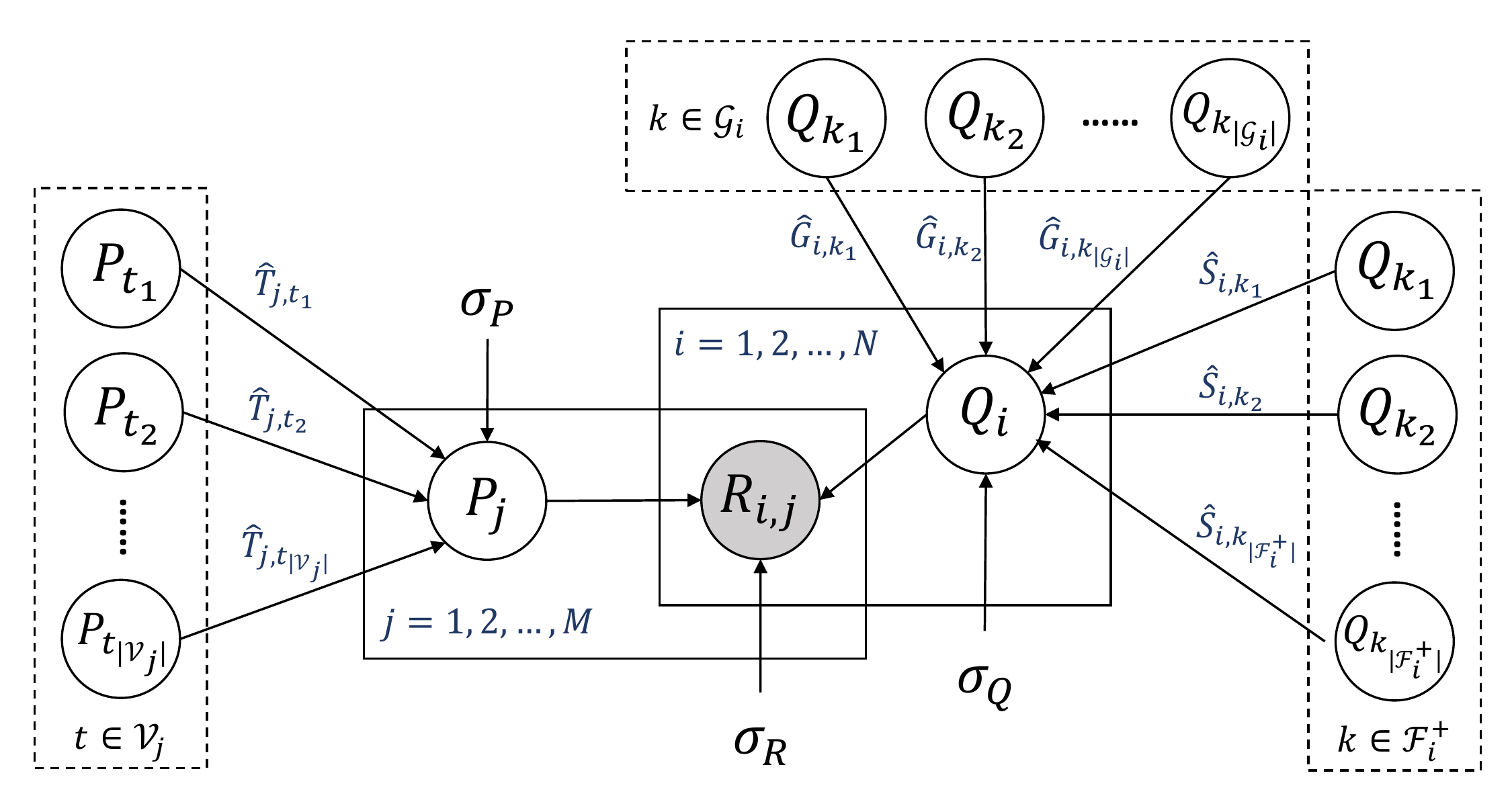}
		\caption{Graphic Model of JTS-MF.}
		\label{fig:graphic_model}\vspace{-0.15in}
	\end{figure}

	In this section, we present our \textit{Joint Topic-Semantic-aware Matrix Factorization} (JTS-MF) model for online social votings, in which social relationship, group affiliation, and topic-semantic similarities are combined and taken into account for voting recommendation in a comprehensive manner.
	Motivated by Locally Linear Embedding \cite{roweis2000nonlinear} which tries to preserve the local linear dependency among inputs in the low-dimensional embedding space, we expect to keep inter-user and inter-voting topic-semantic similarities in latent feature space as well.
	To this end, in JTS-MF model, while the rating $R_{i, j}$ is factorized as user latent feature $\bm{Q_i}$ and voting latent feature $\bm{P_j}$, we deliberately enforce $\bm{Q_i}$ and $\bm{P_j}$ to be dependent on their social-topic-semantic similar counterparts, respectively.
	The graphic model of JTS-MF model is as shown in Figure \ref{fig:graphic_model}.
	
	\subsection{Similarity Coefficients}
		In order to characterize the influence of inter-user common interests and inter-voting content relevance, we first introduce the following three similarity coefficients:
		\begin{itemize}
			\item Normalized social-level similarity coefficient of users: $\widehat{S}_{i, k}$, where $u_k$ is the social-level friend of $u_i$;
			\item Normalized group-level similarity coefficient of users: $\widehat{G}_{i,k}$, where $u_k$ is the group-level friend of $u_i$;
			\item Normalized similarity coefficient of voting: $\widehat T_{j, t}$, where $v_j$ and $v_t$ are two distinct votings.
		\end{itemize}
		
		Generally speaking, in JTS-MF, the latent feature $\bm{Q_i}$ for user $u_i$ is tied up with the latent feature of his social-level and group-level friends who are weighted through $\widehat{S}_{i, k}$'s and $\widehat{G}_{i,k}$'s.
		Likewise, the latent feature $\bm{P_j}$ for voting $v_{j}$ is tied up with the latent feature of its similar votings, which are weighted through $\widehat T_{j, t}$'s.
		
		\subsubsection{Normalized social-level similarity coefficient of users}
			Social-level similarity coefficient of users is represented by matrix $\bm S^{N \times N}$, which incorporates both social relationship and user-user topic-semantic similarity.
			Specifically, for each $u_i$, the social-level similarity coefficient with respect to $u_k$ is defined as
			\begin{equation}
			\label{eq:social_level_user_sim}
				S_{i, k} =I_{u_i, u_k} \cdot \sqrt{\frac{d_k^- + d}{d_i^+ + d_k^- + d}} \cdot \frac{{\bf e}_{u_i}^{\top}{\bf e}_{u_k}}{\|{\bf e}_{u_{i}}\|_{2}\|{\bf e}_{u_k}\|_{2}},
			\end{equation}
			where $I_{u_i, u_k}$ indicates whether $u_i$ follows $u_k$ as described in Eq. (\ref{eq:uu_relation}), $d_i^+$ is the out-degree of $u_i$ in the social network (i.e., $d_i^+ = |\mathcal F_i^+|$), $d_k^-$ is the in-degree of $u_k$ in the social network (i.e., $d_k^- = |\mathcal F_k^-|$), $d$ is the smoothing constant ($d = 1$ in this paper), and $\frac{{\bf e}_{u_i}^{\top}{\bf e}_{u_k}}{\|{\bf e}_{u_{i}}\|_{2}\|{\bf e}_{u_k}\|_{2}}$ is the topic-semantic similarity between user $u_i$ and user $u_k$ as mentioned in Section \ref{section:tewe}.
			$\sqrt{\frac{d_k^- + d}{d_i^+ + d_k^- + d}}$ incorporates the information of local authority and local hub value to differentiate the importance of different users \cite{ma2008sorec}.
			Essentially, $S_{i, k}$ counts the closeness between two users from both topic-semantic interests and their social influence perspectives.
			
			To avoid the impact of different numbers of followees, we use the normalized social-level similarity coefficient of users in JTS-MF, which is defined as
			\begin{equation}
				\widehat S_{i, k} =\frac{S_{i, k}}{\sum_{k \in \mathcal F_i^+} S_{i, k}},
			\end{equation}
			where $\mathcal F_i^+$ denotes the set of $u_{i}$'s followees in social network.
			
		\subsubsection{Normalized group-level similarity coefficient of users}
			Group-level similarity coefficient of users is represented by matrix $\bm G^{N \times N}$, which actually measures the topic-semantic similarity among users from viewpoint of groups.
			For each $u_i$, the group-level similarity coefficient with respect to $u_k$ is defined as
			\begin{equation}
			\label{eq:group_level_user_sim}
				G_{i, k} = \sum\nolimits_{G \in \mathcal G}I_{u_i, G} \cdot I_{u_k, G}\cdot\frac{{\bf e}_{u_i}^\top {\bf e}_G}{ \|{\bf e}_{u_i}\|_{2}\|{\bf e}_{G}\|_{2}},
			\end{equation}
			where $\mathcal G$ represents the set of all groups, $I_{u_i, G}$ and $I_{u_k, G}$ indicate whether $u_{i}$ and $u_{k}$ join group $G$ respectively as described in Eq. (\ref{eq:ug_relation}), and the last term is the topic-semantic similarity between user $u_{i}$ and group $G$.
			Essentially speaking, $G_{i, k}$ reflects the interest closeness between user $u_{i}$ and its group-level friend $u_{k}$ by using $u_{i}$'s topic-semantic engagement extent to the corresponding group.
			We also normalize the group-level similarity coefficient of users as
			\begin{equation}
				\widehat G_{i, k} =\frac{G_{i, k}}{\sum_{k \in \mathcal G_i} G_{i, k}},
			\end{equation}
			where $\mathcal G_i$ is the set of $u_{i}$'s group-level friends in social network.
		
		\subsubsection{Normalized similarity coefficient of votings}
			Similarity coefficient of votings is represented by matrix $\bm T^{M \times M}$, which is directly defined as the topic-semantic similarity among votings, i.e., 
			\begin{equation}
			\label{eq:voting_sim}
				T_{j, t} = \frac{{\bf e}_{v_j}^\top {\bf e}_{v_t}}{\|{\bf e}_{v_j}\|_{2}\|{\bf e}_{v_t}\|_{2}}.
			\end{equation}
			
			Since the number of votings is typically huge, we only consider the similarity between two votings with sufficiently high coefficient value.
			Specifically, for each voting $v_j$, we define a set of votings $\mathcal V_j$ containing those votings whose similarity coefficients with $v_j$ exceed a threshold, i.e., $\mathcal V_j = \{ v_t | T_{j, t} \geq threshold \}$.
			Correspondingly, the similarity coefficient of votings are normalized as
			\begin{equation}
				\widehat T_{j, t} =\frac{T_{j, t}}{\sum_{t \in \mathcal V_j} T_{j, t}}.
			\end{equation}

	\subsection{Objective Function}
		Using the notations listed above, the objective function of JTS-MF can be written as
		\begin{equation}
			\label{eq:objective_function}
			\small
			\begin{split}
				L & = \frac{1}{2} \sum_{i=1}^N \sum _{j=1}^M I'_{i, j}\left(R_{i, j} - \bm{Q_i} \bm{P_j}^{\top}\right)^2 + \frac{\alpha}{2} \sum_{i=1}^N \big\| \bm Q_i - \sum_{k \in \mathcal F^+_i} \widehat S_{i, k} \bm Q_k \big\|^2_2\\
				& + \frac{\beta}{2} \sum_{i=1}^N \big\| \bm Q_i - \sum_{k \in \mathcal G_{i}} \widehat G_{i, k} \bm Q_k \big\|^2_2 + \frac{\gamma}{2} \sum_{j=1}^M \big\| \bm P_j - \sum_{t \in \mathcal V_j} \widehat T_{j, t}  \bm P_t \big\|^2_2 + \frac{\lambda}{2} \left( \|\bm Q\|_F^2 + \|\bm P\|_F^2 \right).
			\end{split}
		\end{equation}
		
		The basic idea of the objective function in Eq. (\ref{eq:objective_function}) lies in that, besides considering explicit feedback between users and votings, we also impose penalties on the discrepancy among features of similar users and similar votings.
		We give detailed explanation as follows.
		The first term of Eq. (\ref{eq:objective_function}) measures the mean squared error between prediction and ground truth, where $I'_{i, j}$ is the training weights defined as
		\begin{equation}
			\label{eq:im}
			I'_{i, j}=
			\begin{cases}
				1, & if \ u_i \ participates \ v_j\\
				I_m, & otherwise
			\end{cases}.
		\end{equation}
		The reason we do not directly use $I_{u_i, v_j}$ defined in Eq. (\ref{eq:uv_relation}) as the training weights is because we found a small and positive $I_m$ makes the training process more robust and can greatly improve the results.
		$R_{i, j}$ is the actual rating of user $u_i$ on voting $v_j$, and $\bm{Q_i} \bm{P_j}^{\top}$ is the predicted value of $R_{i, j}$.
		Without loss of generality, in JTS-MF model, we set $R_{i, j}=1$ if $u_i$ participates $v_j$ and $R_{i, j}=0$ otherwise.
		
		The second, third, and fourth terms of Eq. (\ref{eq:objective_function}) measure the penalty of discrepancy among similar users and similar votings.
		In particular, the second term enforces user $u_i$'s latent feature $\bm{Q_i}$ to be similar to the weighted average of his like-minded followees' profiles $\bm{Q_k}$'s.
		Weight $\widehat S_{i, k}$'s address both the followee $u_{k}$'s social influence on $u_{i}$ as well as the degree of common voting interests shared between $u_{k}$ and $u_{i}$.
		The third term enables user $u_i$'s latent feature $\bm{Q_i}$ to be similar to the weighted average of all his group peers' profiles $\bm{Q_k}$'s.
		Weight $\widehat G_{i, k}$'s emphasize both the same group affiliation of users $u_{i}$ and $u_{k}$ and also the tie strength between $u_{i}$ and the associated group with respect to voting interests.
		This implies that, among all group-level friends, $u_{i}$ would have more similar latent feature with the users who frequently join those groups $u_{i}$ is interested in.
		The fourth term ensures voting $v_j$'s latent feature $\bm{P_j}$ to be similar to the weighted average of votings that share similar topic-semantic information with $v_j$.
		
		Finally, the last term of Eq. (\ref{eq:objective_function}) is the regularizer to prevent over-fitting, and $\lambda$ is the regularization weight.
			
		The trade-off among user social-level similarities, user group-level similarities, and voting similarities is controlled by the parameters $\alpha$, $\beta$, and $\gamma$, respectively.
		Obviously, users' social-level similarity, users' group-level similarity, or votings' similarity is/are ignored if $\alpha$, $\beta$, or $\gamma$ is/are set to 0, while increasing these values shifts the trade-off more towards their respective directions.

		\subsection{Learning Algorithm}
			To solve the optimization in Eq. (\ref{eq:objective_function}), we apply batch gradient descent approach to minimize the objective function\footnote{Note that it is impractical to apply Alternating Least Squares (ALS) method here because it requires calculating the inverse of two matrices with extremely large size.}.
			The gradients of loss function in Eq. (\ref{eq:objective_function}) with respect to each variable $\bm{Q_i}$ and $\bm{P_j}$ are as follows:
			\begin{equation}
			\label{eq:gradients_1}
			\small
				\begin{split}
					\frac{\partial L}{\partial \bm{Q_i}} &= \sum _{j=1}^M - I'_{i, j}\left(R_{i, j} - \bm {Q_i} \bm{P_j}^{\top} \right) \bm{P_j}\\
					&+ \alpha \bigg( \big( \bm {Q_i} - \sum_{k \in \mathcal F^+_i} \widehat S_{i, k}  \bm {Q_k} \big) + \sum_{t \in \mathcal F^-_i} -\widehat S_{t, i} \big( \bm {Q_t} - \sum_{k \in \mathcal F^+_t} \widehat S_{t, k}  \bm {Q_k} \big) \bigg)\\
					&+ \beta \bigg( \big( \bm {Q_i} - \sum_{k \in \mathcal G_{i}} \widehat G_{i, k}  \bm {Q_k} \big) + \sum_{t \in \mathcal U} -\widehat G_{t, i} \big( \bm {Q_t} - \sum_{k \in \mathcal G_{i}} \widehat G_{t, k}  \bm {Q_k} \big) \bigg)
					+ \lambda \bm {Q_i},
				\end{split}
			\end{equation}
			\begin{equation}
			\label{eq:gradients_2}
			\small
				\begin{split}
					\frac{\partial L}{\partial \bm{P_j}} &= \sum _{i=1}^N - I'_{i, j}\left(R_{i, j} - \bm {Q_i} \bm{P_j}^{\top} \right) \bm{Q_i}\\
					&+ \gamma \bigg( \big( \bm {P_j} - \sum_{t \in \mathcal V_j} \widehat T_{j, t}  \bm {P_t} \big) + \sum_{k \in \mathcal V_j} -\widehat T_{k, j} \big( \bm {P_k} - \sum_{t \in \mathcal V_k} \widehat T_{k, t}  \bm {P_t} \big) \bigg)
					+ \lambda \bm {P_j}.
				\end{split}
			\end{equation}
			
			To clearly understand the gradients in Eq. (\ref{eq:gradients_1}) and (\ref{eq:gradients_2}), it is worth pointing out that $\bm{Q_i}$ appears not only in the $i$-th sub-term in the second and third lines of Eq. (\ref{eq:objective_function}) explicitly, but also exists in other $t$-th sub-terms followed by $\widehat S_{t, i}$ or $\widehat G_{t, i}$, where $u_{i}$ plays as one of the followees or group members of other users.
			The case is similar for $\bm{P_j}$.
			Given the gradients in Eq. (\ref{eq:gradients_1}) and (\ref{eq:gradients_2}), we list the pseudo code of the learning algorithm for JTS-MF as follows:
			\begin{enumerate}
				\item
					Randomly initialize $\bm Q$ and $\bm P$;
				\item
					In each iteration of the algorithm, do:\\
					a) update each $\bm {Q_i}$: $\bm{Q_i} \leftarrow \bm{Q_i} - \delta \frac{\partial L}{\partial \bm{Q_i}}$;\\
					b) update each $\bm {P_j}$: $\bm{P_j} \leftarrow \bm{P_j} - \delta \frac{\partial L}{\partial \bm{P_j}}$;\\
					until convergence, where $\delta$ is an configurable learning rate.
			\end{enumerate}

\section{Experiments}
\label{experiments}
	In this section, we evaluate our proposed JTS-MF model on the aforementioned Weibo voting dataset\footnote{Experiment code is provided at \url{https://github.com/hwwang55/JTS-MF}.}.
	We first introduce baselines and parameter settings used in the experiments, and then present the experimental results of JTS-MF and the comparison with baselines.                
		
	\subsection{Baselines}
	\label{section:baselines}
		We use the following seven methods as the baselines against JTS-MF model.
		Note that the first three baselines are reduced versions of JTS-MF, which only consider one particular type of similarity among users or votings.
		
		\begin{itemize}
			\item
				\textbf{JTS-MF(S)} only considers social-level similarity of users, i.e., sets $\beta, \gamma = 0$ in JTS-MF model.
			\item
				\textbf{JTS-MF(G)} only considers group-level similarity of users, i.e., sets $\alpha, \gamma = 0$ in JTS-MF model.
			\item
				\textbf{JTS-MF(V)} only considers similarity of votings, i.e., sets $\alpha, \beta = 0$ in JTS-MF model.
			\item
				\textbf{MostPop} recommends the most popular items to users, i.e., the votings that have been participated by the most numbers of users.
			\item
				\textbf{Basic-MF} \cite{koren2009matrix} simply uses matrix factorization method to predict the user-voting matrix while ignores additional social relation, group affiliation and voting content information.
			\item
				\textbf{Topic-MF} \cite{Blei2003LDA} is similar to JTS-MF except that we substitute $\Theta_d$ for ${\bf e}_d$ when calculating similarities in Eq. (\ref{eq:social_level_user_sim}), (\ref{eq:group_level_user_sim}), and (\ref{eq:voting_sim}).
				Note that $\Theta_d$ can also be viewed as the embedding of document with respect to topics.
				Therefore, Topic-MF only considers the topic similarity among users and votings.
			\item
				\textbf{Semantic-MF} is similar to JTS-MF except that we use the Skip-Gram model in \cite{Mikolov2013wordembedding} directly to learn the word embeddings.
				Therefore, Semantic-MF only considers the semantic similarity among users and votings.
        	\end{itemize}

	\subsection{Parameter Settings}
		We use GibbsLDA++\footnote{GibbsLDA++: \url{http://gibbslda.sourceforge.net}}, an open-source implementation of LDA using Gibbs sampling, to calculate topic information of words and documents in JTS-MF and Topic-MF models.
		We set the number of topics to 50 and leave all other parameters in LDA as default values.
		For word embeddings in JTS-MF and Semantic-MF models, we use the same settings as follows: length of embedding dimension as 50, window size as 5, and number of negative samples as 3.
		
		For all MF-based methods, we set the learning rate $\delta = 0.001$ and regularization weight $\lambda = 0.5$ by 10-fold cross validation.
		Typically, we set $I_m = 0.01$ in Eq. (\ref{eq:im}).
		Taking into consideration the balance of experimental results and time complexity, we run 200 iterations for each of the experiment cases.
		To conduct the recommendation task, we randomly select 20\% of users' voting records in the dataset as test set and use the remaining data as the trainning examples for our JTS-MF model as well as all baselines.
		The choice of remaining hyper-parameters (trade-off parameters $\alpha$, $\beta$, $\gamma$, and dimension of latent features $dim$) is discussed in Section \ref{section:parameter_sensitivity}.
		
		To quantitatively analyze the performance of voting recommendation, in our experiment, we use \textit{top-k recall} (\textit{Recall@k}), \textit{top-k precision} (\textit{Precision@k}), and \textit{top-k micro-F1} (\textit{Micro-F1@k}) as the evaluation metrics.

	\subsection{Experiment Results}
		\subsubsection{Study of convergence}
		\label{section:study_of_convergence}
\begin{figure}
				\centering
				\includegraphics[width=.35\textwidth]{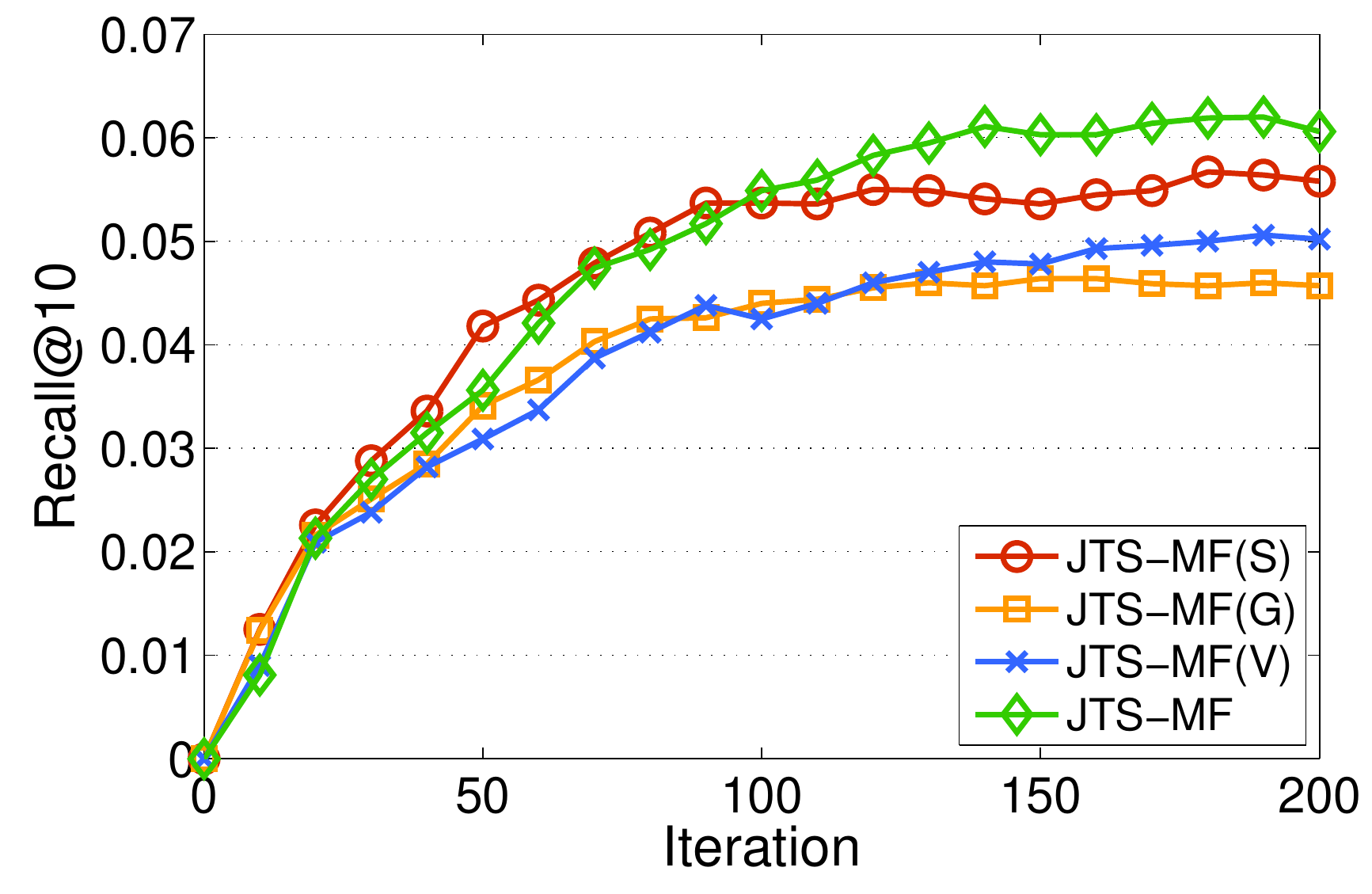}\vspace{0.05in}
				\caption{Convergence of JTS-MF models with respect to \textit{Recall@10}.}
				\label{fig:iteration}
			\end{figure}

			To study the convergence of JTS-MF model, we run the learning algorithm up to 200 iterations for JTS-MF(S) with $\alpha=10$, JTS-MF(G) with $\beta=140$, JTS-MF(V) with $\gamma=30$, JTS-MF with $\alpha=10$, $\beta=140$, $\gamma=30$ ($dim=10$ for $\bm{Q_i}$ and $\bm{P_j}$ in all models), then calculate \textit{Recall@10} for every 10 iterations.
			The result of convergence of JTS-MF models is plotted in Fig. \ref{fig:iteration}.
			From Fig. \ref{fig:iteration} we can see that, the recall of JTS-MF models rises rapidly before 100 iterations, and starts to oscillate slightly after around 150 iterations.
			The same changing pattern is observed for all four JTS-MF variants.
			Therefore, we set the number of learning iterations as 200 to achieve a balance between running time and performance of models.
			
		\subsubsection{Study of JTS-MF}
\begin{figure*}
				\centering
            			\begin{subfigure}[b]{0.3\textwidth}
                			\includegraphics[width=\textwidth]{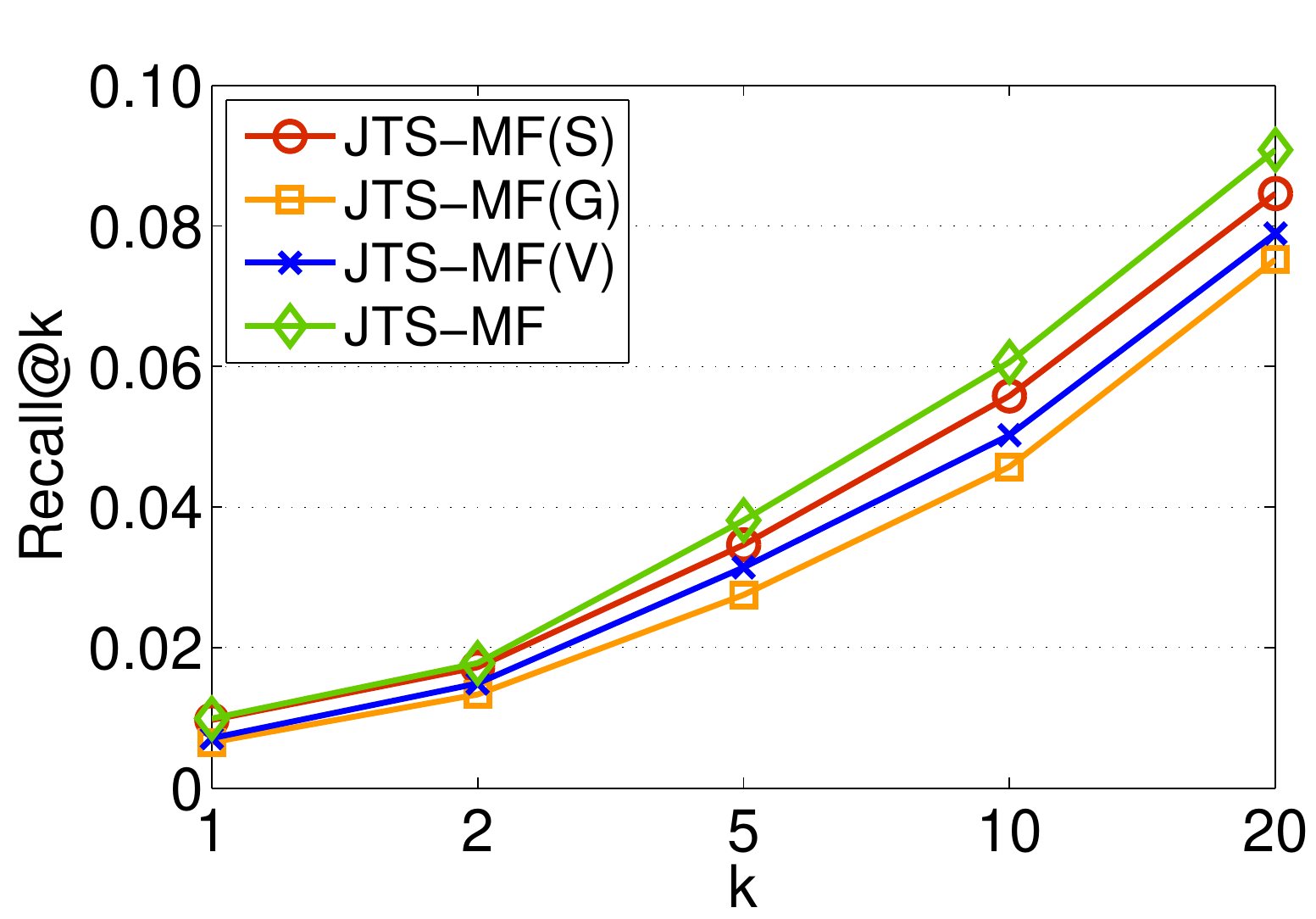}
                			\caption{}
                			\label{fig:recall}
            			\end{subfigure}
            			\hfill
            			\begin{subfigure}[b]{0.3\textwidth}
                			\includegraphics[width=\textwidth]{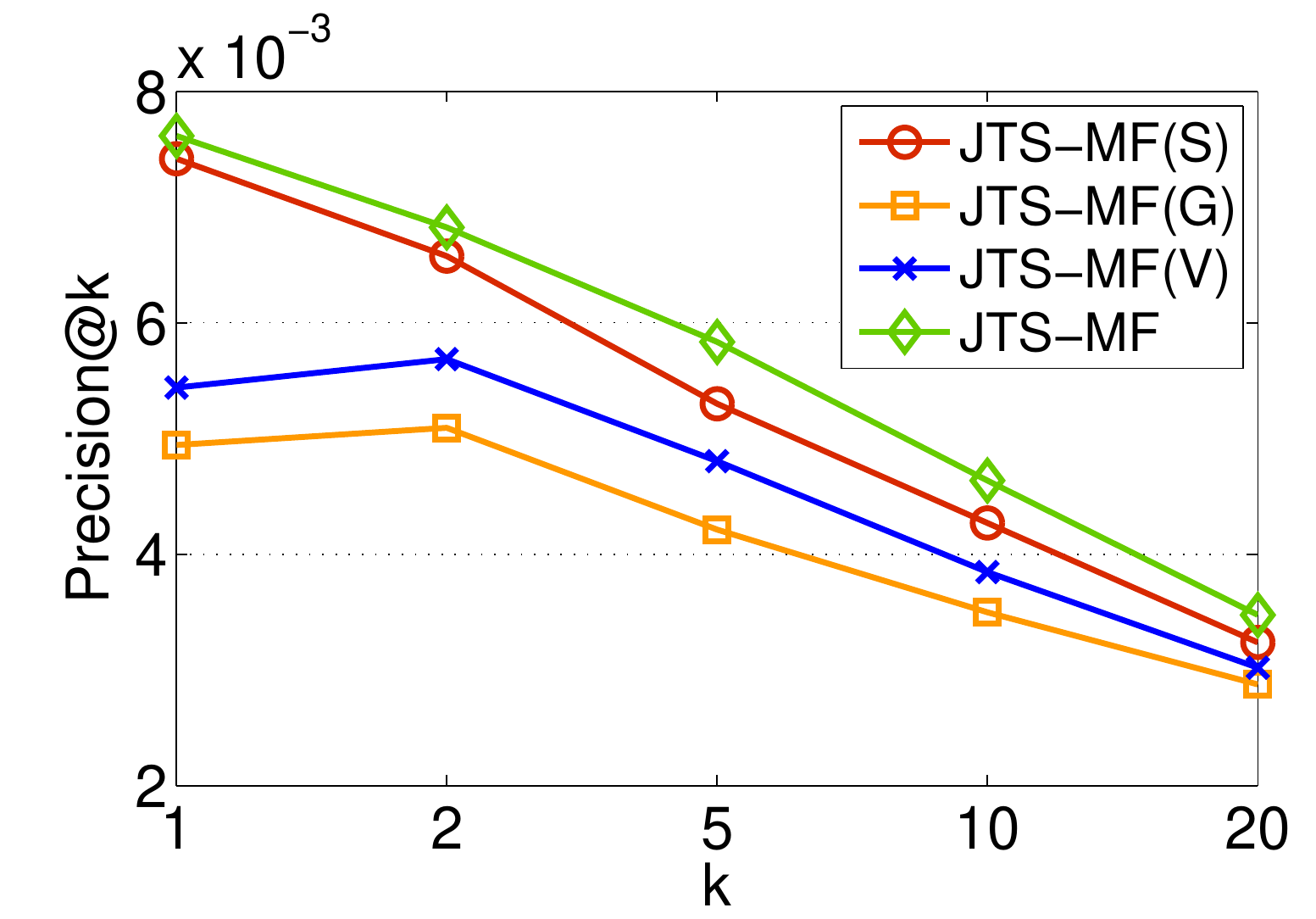}
                			\caption{}
                			\label{fig:precision}
            			\end{subfigure}
            			\hfill
            			\begin{subfigure}[b]{0.3\textwidth}
                			\includegraphics[width=\textwidth]{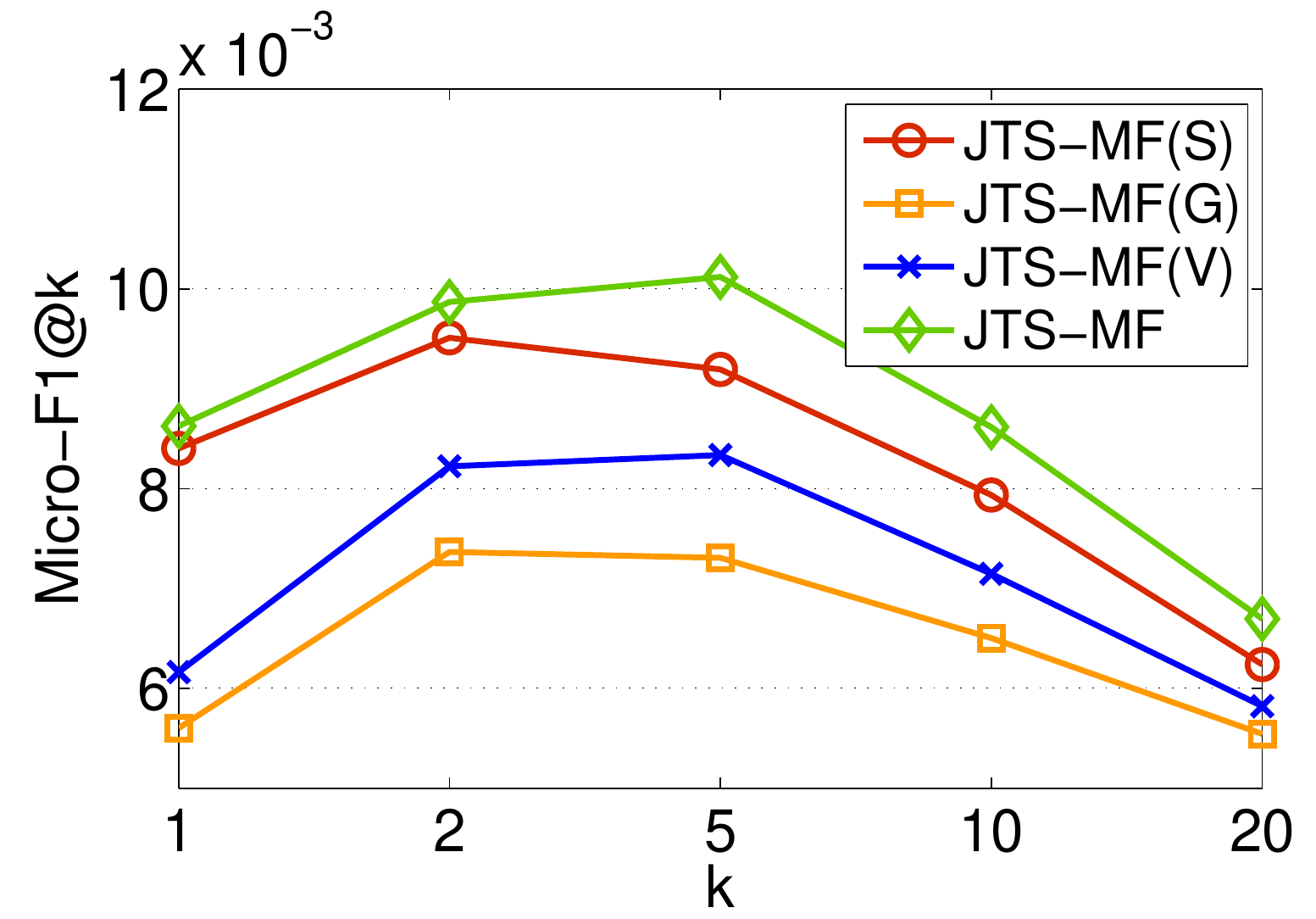}
                			\caption{}
                			\label{fig:microf1}
            			\end{subfigure}
            			\vspace{0.05in}
            			\caption{(a) \textit{Recall@k}, (b) \textit{Precision@k}, and (c) \textit{Micro-F1@k} of JTS-MF models.}
            			\label{fig:jts-mf}
        		\end{figure*}

			To study the performance of JTS-MF model and the effectiveness of three types of similarities, we run JTS-MF model as well as its three reduced versions on Weibo voting dataset, and report the results of \textit{Recall}, \textit{Precision}, and \textit{Micro-F1} in Fig. \ref{fig:jts-mf}. The parameter settings of $\alpha$, $\beta$, $\gamma$, and $dim$ are the same as in Section \ref{section:study_of_convergence}.
			Fig. \ref{fig:recall}, \ref{fig:precision}, and \ref{fig:microf1} consistently demonstrate that JTS-MF(S) performs best and JTS-MF(G) performs worst among three types of reduced versions of JTS-MF.
			Note that JTS-MF(S) only considers users' social-level similarity and JTS-MF(G) only considers users' group-level similarity.
			Therefore, it could be concluded that social-level friends are more helpful than group-level friends when determining users' voting interest.
			This is in accordance with our intuition, since a user typically has much more group-level friends than social-level friends, which inevitably dilutes its effectiveness and brings noises into group-level relationship.
			In addition, the result in Fig. \ref{fig:jts-mf} also demonstrates the effectiveness of the usage of votings' similarity.
			Furthermore, it can be evidently observed that JTS-MF model outperforms its three reduced versions in all cases, which proves that the three types of similarities are well combined in JTS-MF model to achieve much better results.
			
        	\begin{figure*}
			\centering
            		\begin{subfigure}[b]{0.24\textwidth}
                		\includegraphics[width=\textwidth]{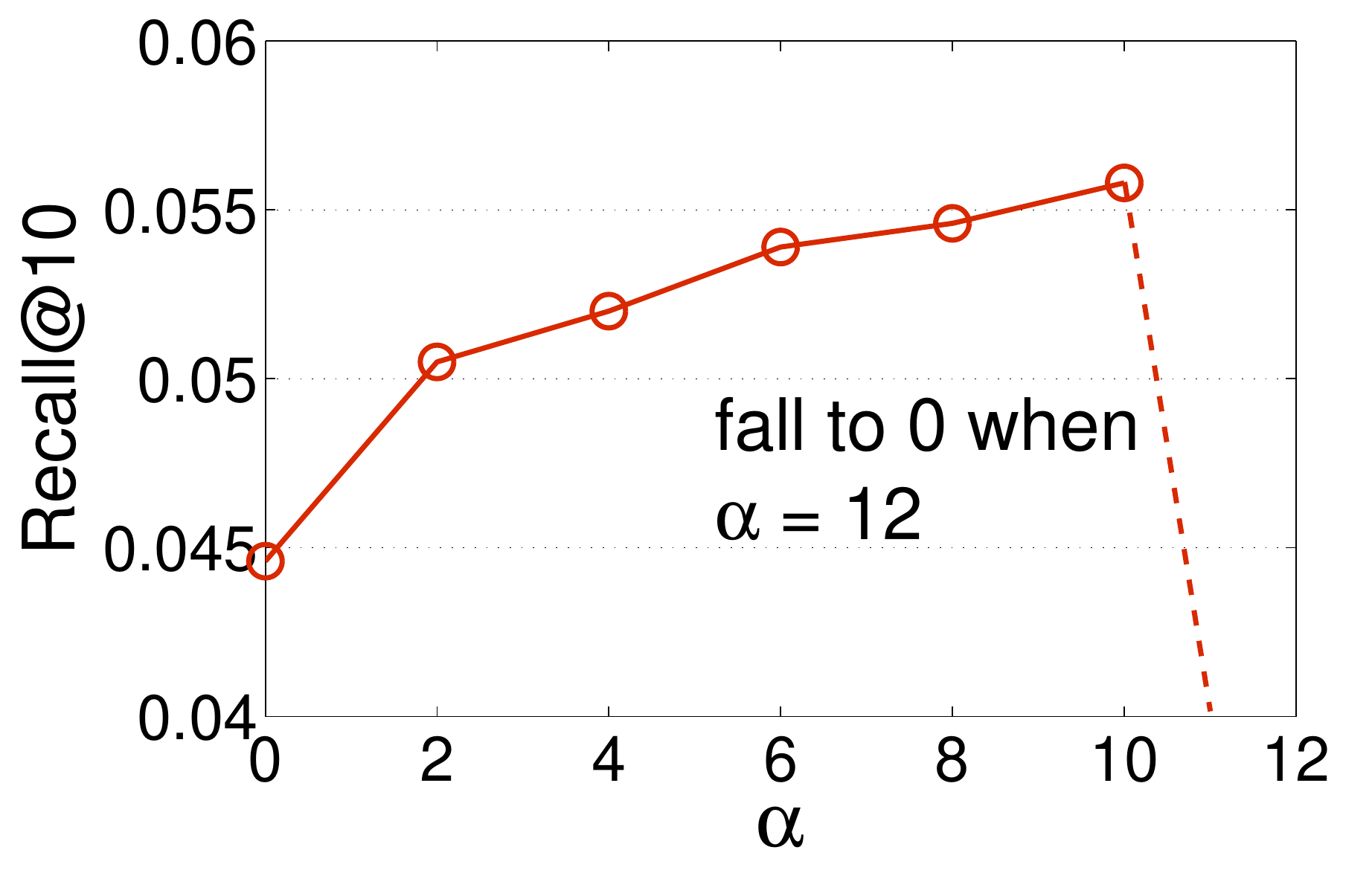}
                		\caption{}
                		\label{fig:alpha}
            		\end{subfigure}
            		\hfill
            		\begin{subfigure}[b]{0.24\textwidth}
                		\includegraphics[width=\textwidth]{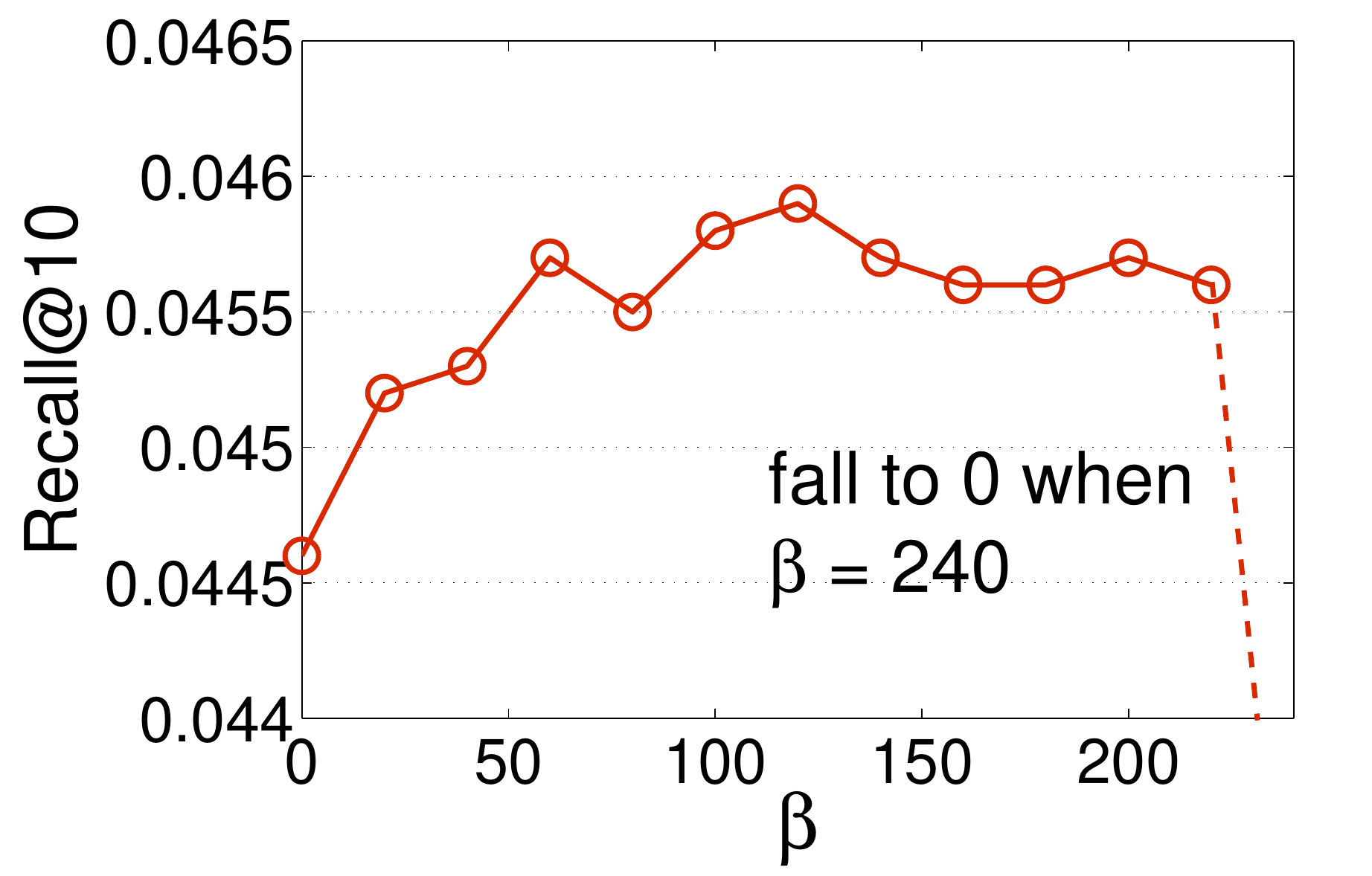}
                		\caption{}
                		\label{fig:beta}
            		\end{subfigure}
            		\hfill
            		\begin{subfigure}[b]{0.24\textwidth}
                		\includegraphics[width=\textwidth]{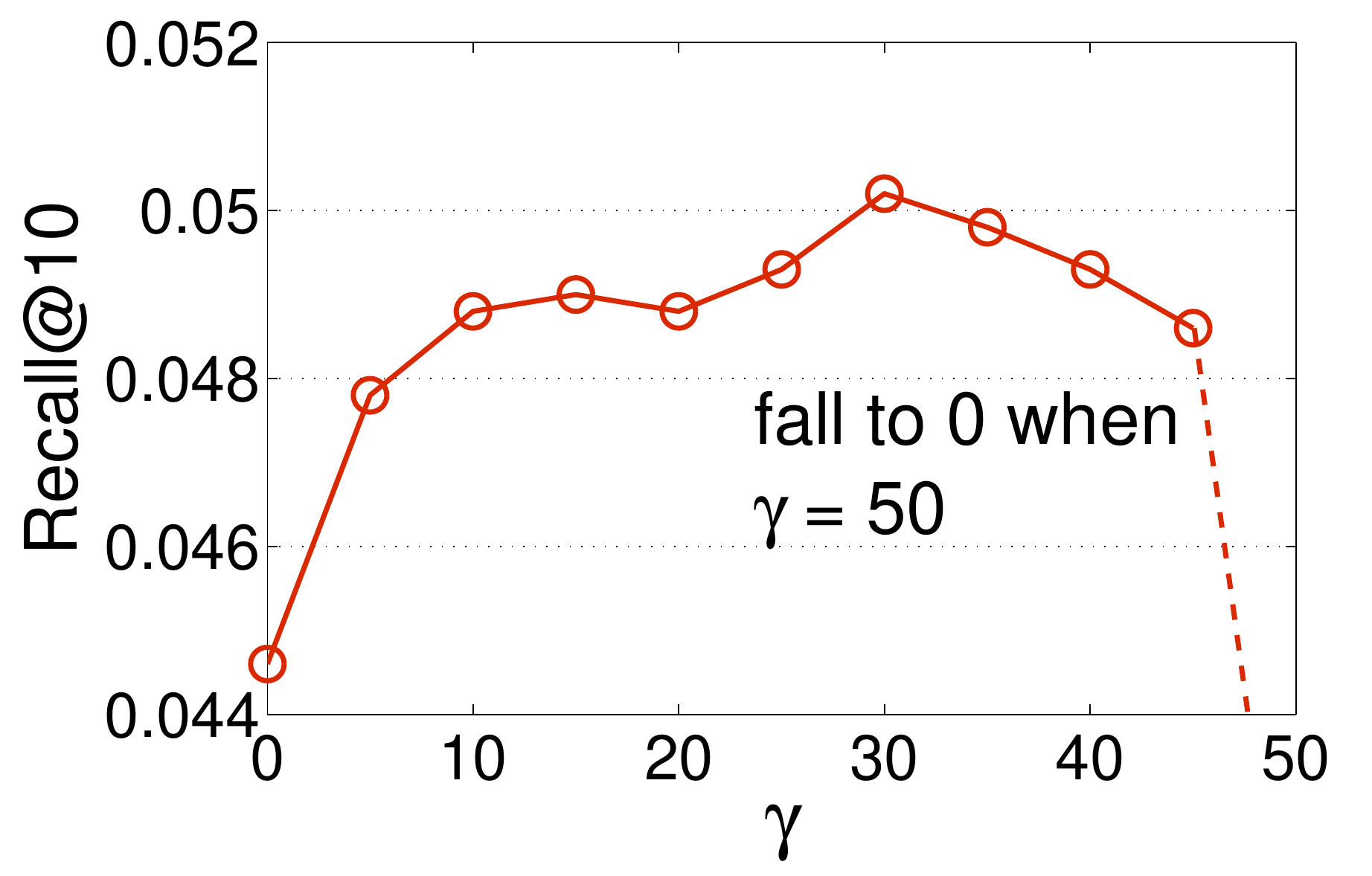}
                		\caption{}
                		\label{fig:gamma}
            		\end{subfigure}
            		\begin{subfigure}[b]{0.24\textwidth}
                		\includegraphics[width=\textwidth]{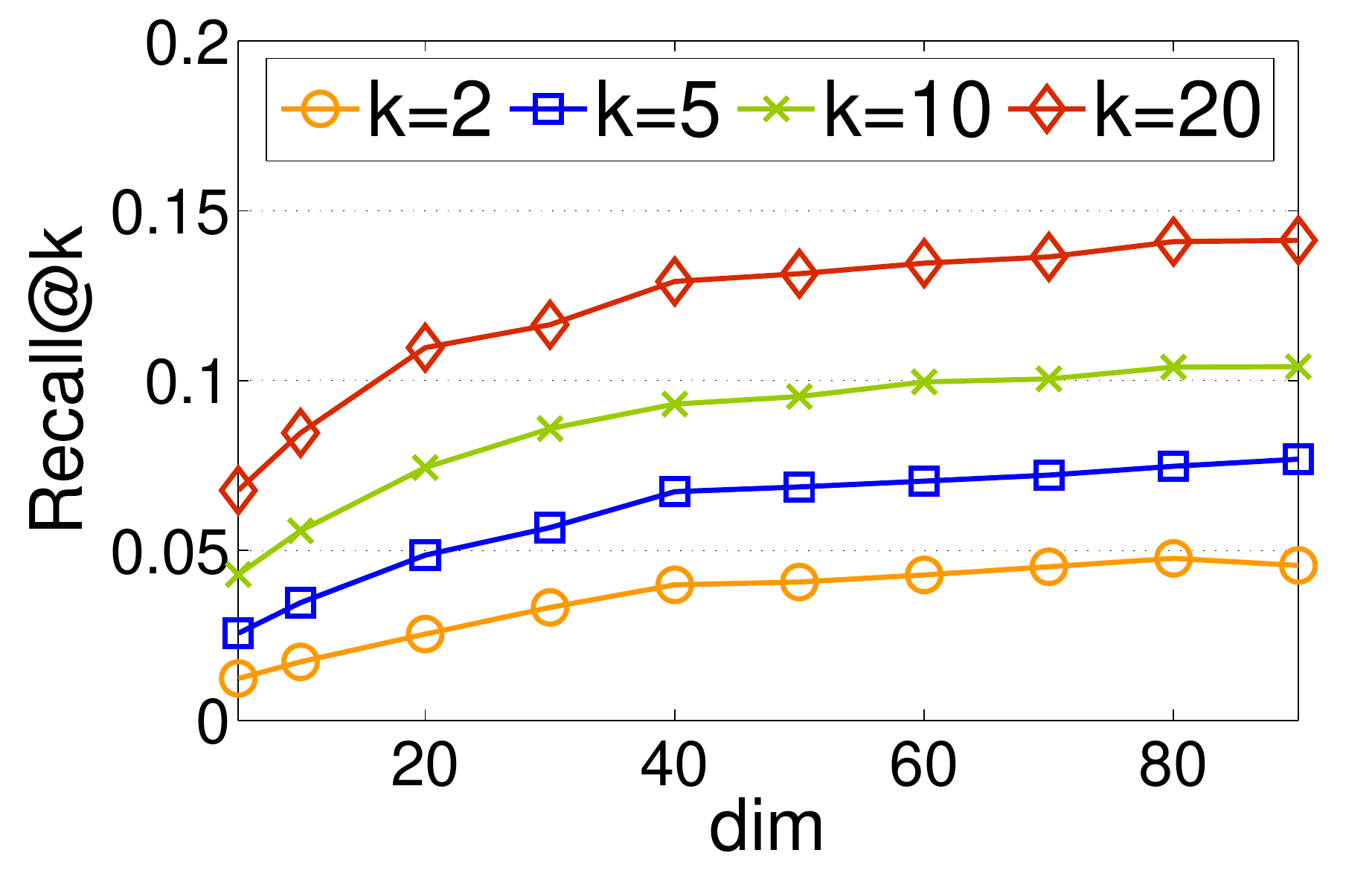}
                		\caption{}
                		\label{fig:dim}
            		\end{subfigure}
            		\vspace{0.05in}
            		\caption{Parameter sensitivity with respect to (a) $\alpha$, (b) $\beta$, (c) $\gamma$, and (d) $dim$.}
            		\label{fig:parameter_sensitivity}
        	\end{figure*}
		
		\subsubsection{Comparison of Models}
			To further compare JTS-MF model with other baselines, we gradually increase $k$ from $1$ to $500$ and report the results in Table \ref{table:all_models} with the best performance highlighted in bold.
			The value of $\alpha$, $\beta$, and $\gamma$ for JTS-MF and its reduced models are the same as in Section \ref{section:study_of_convergence}.
			The parameter settings are $\alpha=2$, $\beta=60$, $\gamma=15$ for Topic-MF, $\alpha=8$, $\beta=120$, $\gamma=20$ for Semantic-MF, and $dim=10$ for $\bm{Q_i}$ and $\bm{P_j}$ in all MF-based methods.
			The above parameter settings are the optimal results of fine tuning for given $dim$.
			In Table \ref{table:all_models}, several observations stand out:
			
			\begin{itemize}
			\item
				MostPop performs worst among all methods, because MostPop simply recommends the most popular votings to all users without considering users' specific interests.
			\item
				Topic-MF and Semantic-MF outperforms Basic-MF, which proves the usage of similarities with respect to topic and semantic helpful for recommending votings.
				Besides, Semantic-MF outperforms Topic-MF.
				This suggests that semantic information is more accurate than topic information when measuring similarities through mining short-length texts.
			\item
				JTS-MF outperforms Topic-MF and Semantic-MF.
				This is the most important observation from Table \ref{table:all_models}, since it justifies our aforementioned claim that joint-topic-semantic model can benefit from both topic and semantic aspects and achieve better performance.
			\item
				The significance of JTS-MF over other models is evident for small $k$.
				However, this margin becomes smaller when $k$ gets larger, and JTS-MF is even slightly inferior to JTS-MF(S) when $k\geq 50$.
				This means that users' group-level similarities and votings' similarities ``drag the feet'' of JTS-MF model when $k$ is large.
				However, JTS-MF is still preferred in practice, since a real recommender system would only recommend a small set of votings to users in general.
        	\end{itemize}

		\begin{table*}
			\small
			\centering
			\caption{Result of \textit{Recall@k}, \textit{Precision@k}, and \textit{Micro-F1@k} for JTS-MF model and baselines.}
			\begin{tabular}{|c|c|c|c|c|c|c|c|c|c|}
				\hline
				\multirow{2}{*}{Model} & \multirow{2}{*}{Metric} & \multicolumn{8}{c|}{$k$}\\
				\cline{3-10}
				& & 1 & 2 & 5 & 10 & 20 & 50 & 100 & 500\\
				\hline
				& \textit{Recall} & 0.0097 & 0.0172 & 0.0346 & 0.0558 & 0.0846 & \textbf{0.1529} & \textbf{0.2229} & \textbf{0.4392}\\
				\hhline{|~|---------}
				\rowcolor[gray]{0.9}
				\cellcolor{white} JTS-MF(S) & \textit{Precision} & 0.007416 & 0.006575 & 0.005300 & 0.004271 & 0.003238 & \textbf{0.002341} & \textbf{0.001707} & \textbf{0.000672}\\
				\hhline{|~|---------}
				\rowcolor[gray]{0.8}
				\cellcolor{white} & \textit{Micro-F1} & 0.008401 & 0.009511 & 0.009192 & 0.007935 & 0.006238 & \textbf{0.004612} & \textbf{0.003387} & \textbf{0.001343}\\
				\hline
				& \textit{Recall} & 0.0065 & 0.0133 & 0.0275 & 0.0457 & 0.0752 & 0.1360 & 0.2051 & 0.4216\\
				\hhline{|~|---------}
				\rowcolor[gray]{0.9}
				\cellcolor{white} JTS-MF(G) & \textit{Precision} & 0.004944 & 0.005092 & 0.004212 & 0.003500 & 0.002877 & 0.002082 & 0.001570 & 0.000645\\
				\hhline{|~|---------}
				\rowcolor[gray]{0.8}
				\cellcolor{white} & \textit{Micro-F1} & 0.005601 & 0.007365 & 0.007306 & 0.006503 & 0.005542 & 0.004102 & 0.003116 & 0.001289\\
				\hline
				& \textit{Recall} & 0.0071 & 0.0149 & 0.0314 & 0.0502 & 0.0789 & 0.1387 & 0.2049 & 0.4176\\
				\hhline{|~|---------}
				\rowcolor[gray]{0.9}
				\cellcolor{white} JTS-MF(V) & \textit{Precision} & 0.005439 & 0.005685 & 0.004805 & 0.003846 & 0.003021 & 0.002124 & 0.001568 & 0.000639\\
				\hhline{|~|---------}
				\rowcolor[gray]{0.8}
				\cellcolor{white} & \textit{Micro-F1} & 0.006161 & 0.008223 & 0.008335 & 0.007145 & 0.005819 & 0.004184 & 0.003112 & 0.001277\\
				\hline
				& \textit{Recall} & \textbf{0.0099} & \textbf{0.0178} & \textbf{0.0381} & \textbf{0.0606} & \textbf{0.0908} & 0.1520 & 0.2187 & 0.4297\\
				\hhline{|~|---------}
				\rowcolor[gray]{0.9}
				\cellcolor{white} JTS-MF & \textit{Precision} & \textbf{0.007614} & \textbf{0.006823} & \textbf{0.005834} & \textbf{0.004637} & \textbf{0.003475} & 0.002327 & 0.001674 & 0.000658\\
				\hhline{|~|---------}
				\rowcolor[gray]{0.8}
				\cellcolor{white} & \textit{Micro-F1} & \textbf{0.008625} & \textbf{0.009868} & \textbf{0.010118} & \textbf{0.008615} & \textbf{0.006695} & 0.004585 & 0.003322 & 0.001314\\
				\hline
				& \textit{Recall} & 0.0042 & 0.0085 & 0.0191 & 0.0313 & 0.0517 & 0.0974 & 0.1455 & 0.3086\\
				\hhline{|~|---------}
				\rowcolor[gray]{0.9}
				\cellcolor{white} MostPop & \textit{Precision} & 0.003221 & 0.003261 & 0.002921 & 0.002403 & 0.001972 & 0.001482 & 0.001119 & 0.000469\\
				\hhline{|~|---------}
				\rowcolor[gray]{0.8}
				\cellcolor{white} & \textit{Micro-F1} & 0.003637 & 0.004721 & 0.005062 & 0.004468 & 0.003804 & 0.002925 & 0.002218 & 0.000937\\
				\hline
				& \textit{Recall} & 0.0063 & 0.0129 & 0.0274 & 0.0446 & 0.0727 & 0.1368 & 0.2050 & 0.4198\\
				\hhline{|~|---------}
				\rowcolor[gray]{0.9}
				\cellcolor{white} Basic-MF & \textit{Precision} & 0.004845 & 0.004944 & 0.004192 & 0.003411 & 0.002783 & 0.002094 & 0.001569 & 0.000643\\
				\hhline{|~|---------}
				\rowcolor[gray]{0.8}
				\cellcolor{white} & \textit{Micro-F1} & 0.005489 & 0.007151 & 0.007271 & 0.006337 & 0.005361 & 0.004125 & 0.003114 & 0.001283\\
				\hline
				& \textit{Recall} & 0.0076 & 0.0147 & 0.0311 & 0.0495 & 0.0781 & 0.1395 & 0.2076 & 0.4210\\
				\hhline{|~|---------}
				\rowcolor[gray]{0.9}
				\cellcolor{white} Topic-MF & \textit{Precision} & 0.005834 & 0.005636 & 0.004766 & 0.003787 & 0.002991 & 0.002136 & 0.001589 & 0.000644\\
				\hhline{|~|---------}
				\rowcolor[gray]{0.8}
				\cellcolor{white} & \textit{Micro-F1} & 0.006609 & 0.008152 & 0.008266 & 0.007035 & 0.005761 & 0.004207 & 0.003154 & 0.001287\\
				\hline
				& \textit{Recall} & 0.0093 & 0.0169 & 0.0333 & 0.0545 & 0.0860 & 0.1471 & 0.2142 & 0.4293\\
				\hhline{|~|---------}
				\rowcolor[gray]{0.9}
				\cellcolor{white} Semantic-MF & \textit{Precision} & 0.007120 & 0.006476 & 0.005102 & 0.004173 & 0.003293 & 0.002252 & 0.001639 & 0.000657\\
				\hhline{|~|---------}
				\rowcolor[gray]{0.8}
				\cellcolor{white} & \textit{Micro-F1} & 0.008065 & 0.009368 & 0.008849 & 0.007752 & 0.006342 & 0.004437 & 0.003254 & 0.001313\\
				\hline
			\end{tabular}
			\label{table:all_models}
		\end{table*}

	\subsection{Parameter Sensitivity}
	\label{section:parameter_sensitivity}
		We investigate parameter sensitivity in this subsection.
		Specifically, we evaluate how different value of trade-off parameters $\alpha$, $\beta$, $\gamma$, and different numbers of latent feature dimensions $dim$ can affect the performance.
		
		\subsubsection{Trade-off parameters}
			We fix $dim = 10$, keep two of the trade-off parameters as 0, and vary the value of the left trade-off parameter.
			Then we report \textit{Recall@10} in Fig. \ref{fig:alpha}, \ref{fig:beta}, and \ref{fig:gamma}, respectively.
			
			As shown in Fig. \ref{fig:alpha}, the \textit{Recall@10} increases constantly as $\alpha$ gets larger and reaches a maximum of 0.0558 when $\alpha = 10$.
			This suggests that the usage of users' social-level similarity do help to improve the recommendation performance.
			However, when $\alpha$ is too large ($\alpha = 12$), the learning algorithm of JTS-MF is misled to wrong direction when updating latent features of users and votings, resulting in performance deterioration.
			The similar phenomenon are observed in Fig. \ref{fig:beta} and Fig. \ref{fig:gamma}, too.
			According to the results, when the other two trade-off parameters are set to 0, \textit{Recall@10} reaches the maximum when $\alpha = 10$, $\beta = 140$, and $\gamma = 30$, respectively.
			Therefore, in previous experiments we adopt these optimal settings for JTS-MF(S), JTS-MF(G), and JTS-MF(V), respectively, and use their combination as the parameter settings in JTS-MF.
		
		\subsubsection{Dimension of latent features}
			We fix $\alpha = 10$, $\beta = 0$, $\gamma = 0$ and tune the dimension of latent features of users and votings from 5 to 90.
			The result is shown in Fig. \ref{fig:dim}.
			From the figure, we can see clearly that the recall is increasing when $dim$ gets larger, this is because latent features with larger number of dimensions have more capacity to characterize users and votings.
			But a larger $dim$ leads to more running time in experiments.
			Moreover, we notice that the improvement of performance stagnates after $dim$ reaches 80.
			On balance, we set $dim = 10$ in our experiment scenarios to ensure the experiments can complete within rational time duration.

\section{Conclusions}
	In this paper, we study the problem of recommending online votings to users in social networks.
	We first formalize the voting recommendation problem and justify the motivation of leveraging social structure and voting content information.
	To overcome the limitations of topic models and semantic models when learning representation of voting content, we propose Topic-Enhanced Word Embedding method to jointly consider topics and semantics of words and documents.
	We then propose our Joint-Topic-Semantic-aware social Matrix Factorization model to learn latent features of users and votings based on the social network structure and TEWE representation.
	We conduct extensive experiments to evaluate JTS-MF with Weibo voting dataset.
	The experimental results prove the competitiveness of JTS-MF against other state-of-the-art baselines and demonstrate the efficacy of TEWE representation.

\begin{acks}
	This work was partially sponsored by the National Basic Research 973 Program of China under Grant 2015CB352403, the NSFC Key Grant (No. 61332004), PolyU Project of Strategic Importance 1-ZE26, and HK-PolyU Grant 1-ZVHZ.
\end{acks}

\bibliographystyle{ACM-Reference-Format}
\bibliography{sigproc} 

\end{document}